%% file: main.tex
\pdfoutput=1

\documentclass[11pt]{article}

\usepackage{EMNLP2022}
\usepackage{CJKutf8}
\usepackage{makecell}
\usepackage{hyperref}

\usepackage{times}
\usepackage{latexsym}
\usepackage{enumerate}
\usepackage[skip=2pt,font=footnotesize]{caption}

\usepackage[T1]{fontenc}

\usepackage[utf8]{inputenc}
\usepackage{tikz}
\usepackage{caption}
\usepackage{subcaption}

\usepackage{microtype}

\usepackage{inconsolata}
\usepackage{graphicx}
\usepackage{multirow}
\usepackage{subcaption}
\usepackage{arydshln}

\usepackage[flushleft]{threeparttable}
\usepackage{scalerel,graphicx,xparse}

\NewDocumentCommand\emojione{}{\scalerel*{\includegraphics{fig/corgi.pdf}}{X}}

\title{CORGI-PM \emojione: A Chinese Corpus For Gender Bias Probing and Mitigation}

\newcommand*\samethanks[1][\value{footnote}]{\footnotemark[#1]}

\author{\small Ge Zhang\textsuperscript{1\space 3\space 4} \thanks{\quad The two authors contributed equally to this work.}\space, Yizhi Li\textsuperscript{2} \samethanks \space, Yaoyao Wu\textsuperscript{5}, Linyuan Zhang \textsuperscript{6}, Chenghua Lin \textsuperscript{2} \thanks{\quad Corresponding authors.} \space , Jiayi Geng\textsuperscript{7}, Shi Wang \textsuperscript{3} \samethanks \space, Jie Fu \textsuperscript{1}\\
        \vspace{-1.5mm} \small \textsuperscript{1} Beijing Academy of Artificial Intelligence, China \\ \vspace{-1.5mm} \small \textsuperscript{2} Department of Computer Science, The University of Sheffield, UK \\\vspace{-1.5mm}  \small \textsuperscript{3} Institute of Computing Technology, Chinese Academy of Sciences, China \\ \vspace{-1.5mm} \small \textsuperscript{4} University of Michigan Ann Arbor, USA \\ \vspace{-1.5mm} \small \textsuperscript{5} University of Colorado Boulder, USA \\ \vspace{-1.5mm} \small \textsuperscript{6} Sichuan University, China \\ \vspace{-1.5mm} \small \textsuperscript{7} McGill University, Canada \\ \vspace{-1.5mm} \small
\{yizhi.li, c.lin\}@sheffield.ac.uk\textsuperscript{2}, gezhang@umich.edu\textsuperscript{1}, wangshi@ict.ac.cn\textsuperscript{3}}

\begin{document}
\maketitle
\begin{abstract}

As natural language processing (NLP) for gender bias becomes a significant interdisciplinary topic, the prevalent data-driven techniques, such as large-scale language models, suffer from data inadequacy and biased corpus, especially for languages with insufficient resources, such as Chinese. To this end, we propose a Chinese cOrpus foR Gender bIas Probing and Mitigation (\textbf{CORGI-PM}\footnote{Our code is available at \href{https://github.com/yizhilll/CORGI-PM}{GitHub}}), which contains 32.9k sentences with high-quality labels derived by following an annotation scheme specifically developed for gender bias in the Chinese context. Moreover, we address three challenges for automatic textual gender bias mitigation, which requires the models to detect, classify, and mitigate textual gender bias. We also conduct experiments with state-of-the-art language models to provide baselines. To our best knowledge, CORGI-PM is the first sentence-level Chinese corpus for gender bias probing and mitigation.
\end{abstract}

\section{Introduction}
Increasing recognition in consensus is that identifying and preventing toxic gender attitudes and stereotypes is essential for society \cite{blodgett2020language}. Since gender-biased information could be presented and widely propagated in textual format, it is essential to develop automatic methods for detecting and mitigating textual gender bias.

Natural language processing (NLP) has been widely used in text-related applications, which have a significant influence on gender bias topics~\cite{costa2019analysis}. On the one hand, large-scale language models (LMs), as a key technique of modern NLP, are proven to learn the subjective gender bias in the training corpus or even amplify it \cite{zhao2017men} On the other hand, it becomes increasingly promising to apply cutting-edge NLP techniques for probing and mitigating gender bias.

\input{fig/fitler_pipeline}

Building a high-quality text corpus has been one of the key tangents in improving NLP applications for debiasing gender stereotypes in texts~\cite{sun2019mitigating}. Some researchers introduce \textit{automatic} annotation techniques, such as gender-swapped based methods, to create corpora for gender bias mitigation~\cite{lu2020gender, zhao2018learning, rudinger2018gender}. While it is attractive to build a large corpus without heavy labors, automatic gender-swapped based methods highly depend on the quality of base language models and are prone to creating nonsensical sentences~\cite{sun2019mitigating}. To address this issue, some works devote effort to developing \textit{human-annotated} corpora for gender bias mitigation. However, these corpora  either mainly focus on word- or grammar-level bias~\cite{webster2018mind, zhu2020great, sahai2021predicting, zhou2019examining}, or only 
concern about sexism-related topics \cite{jiang2022swsr, chiril2021nice, chiril2020annotated, parikh2019multi}. 

Moreover, existing works on gender bias exclusively focus on English~\cite{costa2019analysis}, where few datasets exist for other influential languages such as Chinese. (N.B. details of generated gender bias corpus with nonsensical Chinese sentences can be found in Appendix~\ref{Appendix:CaseStudy}). We aim to tackle the aforementioned issues by providing a high-quality Chinese human-annotated corpus for contextual-level gender bias probing and mitigation. 

To this end, we propose the Chinese cOrpus foR Gender bIas Probing and Mitigation (\textbf{CORGI-PM}) dataset, which consists of 32.9k human-annotated sentences, including both gender-biased and non-biased samples. For the initial data collection, we propose an automatic method that builds a potentially gender-biased sentence set from existing large-scale Chinese corpora. Inspired by the metric leveraging language models for gender bias score calculation proposed in~\citet{Bolukbasi2016ManIT, jiao2021gender}, the samples containing words of high gender bias scores are recalled, and then reranked and filtered according to their sentence-level gender-biased probability, as illustrated in Fig.~\ref{retriving_pipeline}. To ensure the quality of our corpus, the annotation scheme is carefully designed, and annotators with qualified educational backgrounds are selected to further label and paraphrase the retrieved sentences.

Additionally, we address three challenges based on CORGI-PM, \textit{i.e.}, gender bias detection, classification, and mitigation, which provide clear definitions and evaluation protocols for NLP tasks in gender bias probing and mitigation. In order to provide referential baselines and benchmarks for our proposed challenges, we conduct random data splitting with balanced labels and implement experiments on cutting-edge language models in zero-shot, in-context learning, and fine-tuning paradigms. We discuss the experimental settings and provide result analysis in \S\ref{sec:chanllenge}. The implementation details can be referred to in Appendix~\ref{Appendix.ImplementationDetail}.

In summary, we provide a well-annotated Chinese corpus for gender bias probing and mitigation, along with clearly defined corresponding challenges. With a properly designed annotation scheme, CORGI-PM provides a corpus of high quality that assists models in detecting gender bias in texts. More importantly, other than the 22.5k human-annotated non-biased samples, all the 5.2k biased sentences in our corpus are further labeled with gender bias subclasses and companies with parallel bias-free versions provided by the annotators. Our codes and dataset will be released for the benefit of the community.

\section{Data Collection}
\subsection{Sample Filtering}
We propose an automatic processing method to recall, rerank, and filter annotation candidates from raw corpora using a two-stage filtering from word-level to sentence-level, as illustrated in Fig.~\ref{retriving_pipeline}. The Chinese sentence samples are mainly screened out from the SlguSet~\cite{zhao2021} and the CCL corpus~\cite{zhan2019CCL}.

To recall gender-biased words or retrieve candidate sentences with gender bias scores, we compare the target word/sentence representations with the \textit{seed direction}, which can be calculated by the subtraction between the word embeddings of \texttt{she} and \texttt{he} ~\cite{Bolukbasi2016ManIT, jiao2021gender}.
We leverage different Chinese LMs including ERNIE \cite{zhang2019ernie}, CBert \cite{cui2020revisiting}, and Chinese word vectors \cite{qiu2018revisiting} to acquire the word-level and sentence-level representations. For word-level filtering, we use the mentioned metric to build a vocabulary of high bias scores and recall sentences containing such words from the raw corpora with exact matches. We compute gender bias scores of the crawled sentences and group them by the gender bias keywords acquired in the previous stage for sentence-level filtering. The final sentences for annotation are then selected according to a specific global threshold gender bias score and an in-group threshold rank. 
The word-level filtering process presented as word clouds can be found in Appendix~\ref{sec:appen_cor_cloud}.

\begin{table}[bt]

\centering
\scalebox{0.75}{
    \begin{tabular}{l:c|ccc}
    \hline
\multicolumn{2}{c|}{\textbf{Sample}} &  \multicolumn{3}{c}{\textbf{Quantity}} \\ \cline{3-5}
    \multicolumn{2}{c|}{\textbf{Category}} & Train  & Valid  & Test \\
    \hline \hline 

    \parbox[t]{2mm}{ \multirow{3}{*}{\rotatebox[origin=c]{90}{\textbf{Biased}}}}& AC  & 1.90k & 235 & 237 \\
     
       & DI &  2.70k & 334 & 337 \\
       & ANB &  2.47k & 306 & 309 \\

    \hline
    \multicolumn{2}{c|}{\textbf{Non-biased}} & 21.4k & 516  & 526 \\
    \hline

    \multicolumn{2}{c|}{Overall} & 30.1k & 1391 & 1409 \\
  
    \hline
    \end{tabular}
}
\caption{
Overall Statistics of the CORGI-PM Dataset. 
The notations, \textbf{AC}, \textbf{DI}, and \textbf{ANB} represent specific bias labels described in \S~\ref{sec:annotaion_schem}.
}
\vspace{-5mm}
\label{overall_dataset_stat}
\end{table}

\begin{table*}[!ht]

\centering
\scalebox{0.65}{
    \begin{tabular}{c|ccc|ccc|ccc}
    \hline
    \multicolumn{1}{c|}{\textbf{Linguistic}} &   \multicolumn{3}{c|}{\textbf{Non-biased}} & \multicolumn{3}{c|}{\textbf{Biased}} & \multicolumn{3}{c}{\textbf{Corrected Biased}} \\ \hline
    \multicolumn{1}{c|}{\textbf{Info.}} &  Train  & Valid  & Test & Train  & Valid  & Test & Train  & Valid  & Test \\ 
    \hline \hline
    Word & 724k  & 18.9k & 17.7k & 228k & 24.8k & 28.3k & 265k & 27.1k & 30.0k\\  
    Dictionary & 574k  & 14.4k & 14.1k & 167k & 18.4k & 20.4k & 191k & 19.9k & 21.5k\\  
    Character &  1,156k & 30.1k & 28.1k & 358k & 39.2k & 44.4k & 417k & 42.8k & 46.9k\\
    Sent. Length & 53.952 & 58.397 & 53.473 & 85.837 & 76.087 & 85.214 & 99.839 & 82.853 & 89.939 \\
    \hline 

    \end{tabular}
}
\caption{Linguistic Characteristics of the Corpus.
\textit{Word, Dictionary, and Character} separately denote the total Chinese word number, total unique Chinese word number, and total character number of the specific categories. The sentence lengths are defined as the number of containing characters.}
\label{corrected_stat}
\end{table*}

\subsection{Annotation Scheme}\label{sec:annotaion_schem}

The annotation scheme is designed for gender bias probing and mitigation. For gender bias probing, the annotators are required to provide the following information given a sentence: whether gender bias exists; if so, how the bias is established. For gender bias mitigation, the corrected non-biased version of the biased sentences is also required. We further describe the annotation scheme details in the following paragraphs.

\noindent \textbf{Existence and Categorization.}

The annotators are required to annotate whether the sentence is gender-biased (\textbf{B}) or non-biased (\textbf{N}) in contextual-level or word-level, and further clarify how the bias is established.
Given that our raw data is collected using gender-related keywords or from gender-related corpus, the samples annotated without gender bias are useful human-annotated negative samples for detecting gender bias. 
To additionally provide information about gender bias categorization, we classify gender bias types into three subtypes : (1) Gender Stereotyped activity and career choices \textbf{(AC)}; (2) Gender Stereotyped descriptions and inductions \textbf{(DI)}; and (3) Expressed gender-stereotyped attitudes, norms and beliefs \textbf{(ANB)}. The classification standard is inspired by \cite{king2021gender} and further summed up into the mentioned subtypes.

\noindent \textbf{Bias Mitigation.}
Annotators are also required to mitigate the gender bias of selected sentences while keeping the original semantic information. 
We also ask our annotators to diversify the expressions if applicable.
The major revision patterns can be summarized as follows: 
(1). \textit{Replace} the gender-specific pronouns with neutral pronouns. 
(2). \textit{Replace} the gender-specific adjectives with neutral descriptions with similar semantics definitions. 
(3). \textit{Add} additional comments to neutralize the sentences which cannot be directly mitigated.

\begin{table*}[bt]

\centering
\scalebox{0.7}{
    \begin{tabular}{c|c|ccc:c:ccc:c|cc:c:cc:c}
    \hline
   \multirow{2}{*}{\textbf{Model}} & \multirow{2}{*}{\textbf{Metrics}} &  \multicolumn{4}{c}{\textbf{Classification (Val.)}} & \multicolumn{4}{c}{\textbf{Classification (Test)}} & \multicolumn{3}{|c}{\textbf{Detection (Val.)}} & \multicolumn{3}{c}{\textbf{Detection (Test.)}}\\ \cline{3-16}
     &  & AC & DI & \multicolumn{1}{c}{ANB} & \multicolumn{1}{c}{Avg.} & AC & DI & \multicolumn{1}{c}{ANB} &  Avg. & N & \multicolumn{1}{c}{B} & \multicolumn{1}{c}{Avg.} & N & \multicolumn{1}{c}{B} & Avg. \\
    \hline \hline 
    \multirow{3}{*}{BERT}  & Precision & .609 & .729 & .533 & .624 & .556 & .615 & .521 & .564 & .699 & .950 & .824 & .742 & .980 & .861 \\ 
     & Recall & .594 & .665 & .543 & .601 & .493 & .652 & .585 & .577 & .971 & .591 & .781 & .985 & .662 & .823  \\ 
     & F1-Score  & .602 & .695 & .538 & .612 & .522 & .633 & .551 & .567 & .813 & .729 & .771 & .846 & .790 & .818 \\ 
    \hline
    \multirow{3}{*}{Electra}  & Precision & .587 & .727 & .544 & .619 & .556 & .630 & .516 & .568 & .691 & .936 & .814 & .745 & .974 & .860  \\ 
     & Recall & .758 & .687 & .386 & .610 & .680 & .685 & .373 & .579 & .961 & .570 & .766 & .983 & .656 & .820 \\ 
     & F1-Score  & .661 & .706 & .451 & .606 & .612 & .656 & .433 & .567  & .804 & .708 & .756 & .848 & .784 & .816 \\ 
    \hline
    \multirow{3}{*}{XLNet}  & Precision & .587 & .696 & .523 & .602 & .544& .589 & .527 & .553  & .713 & .928 & .820 & .772 & .959 & .865 \\ 
     & Recall & .622 & .643 & .495 & .587 & .545 & .614 & .514 & .558 & .953 & .620 & .787 & .968 & .722 & .845 \\ 
     & F1-Score  & .604 & .669 & .509 & .594 & .544 & .601 & .520 & .555   & .816  & .743 & .780 & .859 & .824 & .841 \\ 
     \hline
     \multirow{3}{*}{Curie}  & Precision & .695 & .907 & .010 & .537 & .622 & .887 & .009 & .506  & .763 & .665 & .714 & .635 & .825 & .730 \\ 
     & Recall & .395 & .802 & .375 & .524 & .395 & .804 & .010 & .403 & .576 & .825 & .700 & .975 & .584 & .780 \\ 
     & F1-Score  & .504 & .851 & .019 & .458 & .508 & .852 & .019 & .460 & .656  & .736 & .696 & .769 & .684 & .727 \\ 
    \hline
    \end{tabular}
}
\caption{
Baseline Results for Gender Bias Detection and Classification Tasks. The overall metric refers to Marco average. 
The model names and abbreviations refer to \S~\ref{sec:chanllenge_DC}.
Categorical definitions refer to \S~\ref{sec:annotaion_schem}.
}
\label{result_detection_classification}
\end{table*}

\subsection{Corpus Analysis}
\label{sec:corpusAnalysis}

In this section, we report the linguistic statistics of CORGI-PM as Tab.~\ref{overall_dataset_stat}. %
We design a balanced split to create the valid and test set considering the negative-positive ratio and bias subclass proportion in the global distribution. As revealed in Tab.~\ref{corrected_stat}\footnote{We use the \href{https://github.com/fxsjy/jieba}{\texttt{Jieba}} to parse.}, we observe two major differences compared the debiased samples with the original ones: longer and more diverse expressions (N.B. sentence length and vocabulary size of Tab.~\ref{corrected_stat}).  We hypothesize that it is due to human annotators' intention to keep the semantic information unchanged and the sentence coherent while mitigating gender bias. They may use more conjunctions and longer descriptions compared to some gender-biased inherent expressions. More details for quality managing and control can be referred to Appendix~\ref{sec:appen_cor_cloud} and~\ref{sec:appen_qm}.

\section{Gender Bias Mitigation Challenges}\label{sec:chanllenge}

To provide a clear definition for automatic textual gender bias probing and mitigation tasks, we propose corresponding challenges and standardize the evaluation protocols. We address two tasks, detection, and classification, for gender bias probing and formalize the gender mitigation challenge as a text mitigation task.

 \begin{table*}[bt]
 \large
 
\centering{
\scalebox{0.7}{
     \begin{tabular}{c|c|c|ccc|cc}
     \hline
   \multirow{2}{*}{\diaghead{x}{Metrics}{Models}}& \multirow{2}{*}{\textbf{BLEU}} & \multirow{2}{*}{\textbf{METEOR}}  &  \multicolumn{3}{c|}{\textbf{ROUGE-L}} & \multicolumn{2}{c}{\textbf{Human Evaluation}}  \\ 
   &  &  & Recall & Precision & F1 & Coherence & Gender Bias \\ 
     \hline \hline 
     *Davinci & .776  & .879  & .203 & .211 & .205 & 5.25 & 0.96  \\ 
   \hline 
     Ada & .288  & .429  & .407 & .180 & .250  & 5.98 & 1.13 \\ 
     Babbage & .359  & .504  & .716 & .310 & .432 & 6.32 & 0.69 \\  
     Curie & .364  & .506  & .692 & .316 & .434 & 6.21 & 1.20\\
     \hline
     \end{tabular}
 }
 }
 \caption{Baseline Results for Gender Bias Correction task. Metrics details can be found in Appendix~\ref{Appendix.ImplementationDetail}. * suggests using the model in zero-shot paradigm and the others refers to fine-tune.}
 \vspace{-3mm}
 \label{result_correction}
 \end{table*}

\subsection{Challenges of Detection and Classification}\label{sec:chanllenge_DC}
We regard both the gender bias detection and classification challenges as \textit{supervised classification} tasks and evaluate them with metrics of consensus.

\noindent \textbf{Definition.} The gender bias detection challenge can be regarded as a binary classification task, where the model is required to predict the probability that a given sentence contains gender bias. As described in \S~\ref{sec:annotaion_schem}, biased samples are further categorized into one or more kinds. Therefore, we can address the gender classification challenge as a multi-label classification task. The precision, recall, and F1-score are selected as the main metrics in these two challenges. Class-wise metrics and macro average summarized evaluation are required through both valid and test sets to show the performance of language models.

\noindent \textbf{Baselines.} We finetune Chinese language models from three representative different pretrained paradigms, \textit{i.e.}, Chinese BERT, Electra, and XLNet \citet{cui2020revisiting}, for both the detection and classification tasks by adding an additional dense prediction layer.
\footnote{Pretrained models can be found at the\href{https://github.com/ymcui/HFL-Anthology}{\texttt{HFL Anthology}}.} We also provide GPT-3 \cite{brown2020language} curie's few-shot performance for both the detection and classification tasks.
Baseline results of detection and classification show that the classification task is challenging, and there is room for performance improvement in detecting gender bias in CORGI-PM, as revealed in Tab.~\ref{result_detection_classification}.

 \subsection{Challenge of Mitigation}
\noindent \textbf{Definition.} The gender bias mitigation challenge can be regarded as a natural language generation task, where the model is asked to generate the corrected version of biased sentences with the human-annotated ones as references.  

\noindent \textbf{Baselines.} 
We test the GPT-3 \cite{brown2020language} on CORGI-PM in fine-tune experiment setting with three different parameter scales, which are Ada(350M), Babbage(1.3B), and Curie(6.7B), and Davinci(175B) in zero-shot experiment setting. We only provide zero-shot results for Davinci because it is the only released GPT-3 editing model. More implementation and evaluation details are introduced in Appendix~\ref{Appendix.ImplementationDetail}. 

\noindent \textbf{Discussion.} We provide both human evaluation and automated metrics for evaluation. Tab.~\ref{result_correction} reveals that LMs can learn the annotation pattern of mitigating gender bias, and the zero-shot editing model shows competitive performance. 
The observation that fine-tuned Babbage outperforms much larger zero-shot Davinci in the human evaluation, and ROUGE-L reveals that CORGI-PM has the potential to be used as strong supervision of the gender bias mitigation task.
We notice that Davinci tends to apply more conservative edits compared to fine-tuned models.
As a result, the sentences edited by Davinci keep most of the original sentences and always only change pronouns and adjectives from the original sentences, which benefits precision focusing automatic metrics like BLEU \cite{papineni2002bleu}, and METEOR \cite{agarwal2007meteor}. The performance difference between human evaluation and automatic metrics reveals the writing style difference between human and language models. 

\section{Conclusion}
\vspace{-2mm}
We propose CORGI-PM, the first Chinese human-annotated corpus for both gender bias probing and mitigation.
We also address definitions and evaluation metrics for three challenges based on CORGI-PM and test the performances of state-of-the-art language models.
Our proposed challenges can serve as benchmarks for measuring the ability of language models to detect, classify, and mitigate textual gender bias.
Experiments show that our sentences with fine-grained subclass labels can assist the language models in gender bias probing, whilst our parallel human-written debiased data can serve as strong supervision of the generative language models.
In summary, we imply future work utilizing CORGI-PM would be benefited the topic of NLP for gender bias probing and mitigation.

\section*{Limitations}
There are several major limitations in this research work. 
Due to the high requirement of annotators for annotating gender-biased sentences and correcting such sentences, we only choose annotators with higher education, which may lead to potential cognitive bias. 
In addition, we only conduct limited implementations and experiments of testing widely-used Chinese language models' performance in our new challenges. 
More language models and techniques can be further explored in our challenges.

\section*{Ethics Statement}
We carefully consider the ethical implications during the collection process.
The collection of our corpus CORGI-PM sentences only relies on public available corpora for research purposes. 
We have acknowledged the potential usage of our dataset as well as related privacy issues to the annotators and received confirmations before the annotation was initiated.


\bibliography{anthology,custom}
\bibliographystyle{acl_natbib}

\clearpage

\appendix

\section{Gender Bias Analysis of Chinese Language Models}
\label{sec:appendix}
\subsection{Evaluation Method and Data Sets}

We conduct experiments to explore gender bias contained in widely-used Chinese language models for research and industrial use. 
We employ the method \citet{Bolukbasi2016ManIT,jiao2021gender} proposed to assess gender bias. 
The gender bias score for a word is calculated by $ \vec{w} \cdot (\vec{she} - \vec{he}) $based on its word vector. 
A positive value means the word is more relevant to females, while a negative value means the word is more relevant to males. The higher the absolute value of the gender bias score, the more biased the word indicates.

\citeauthor{srivastava2022beyond} propose a big benchmark containing a dataset specifying the existing Chinese career words.
\citeauthor{zhu2020great} propose AGSS, a manual-created Chinese word-level adjective list containing gender bias.
To measure gender bias contained in the language models, we first calculate gender bias scores of words in the word list provided \cite{srivastava2022beyond, zhu2020great} according to the projection method \citet{Bolukbasi2016ManIT, jiao2021gender}. 
We compare the career and adjective word gender bias score vectors to get the observations of LMs' influence on word-level learned gender bias.
To make the observations more clear, we further apply the sign function to the career and adjective word gender bias score vectors.
The similarity function used for the heatmaps is Pearson similarity.

\begin{figure}[bt]
\centering

\includegraphics[width=0.9\linewidth]{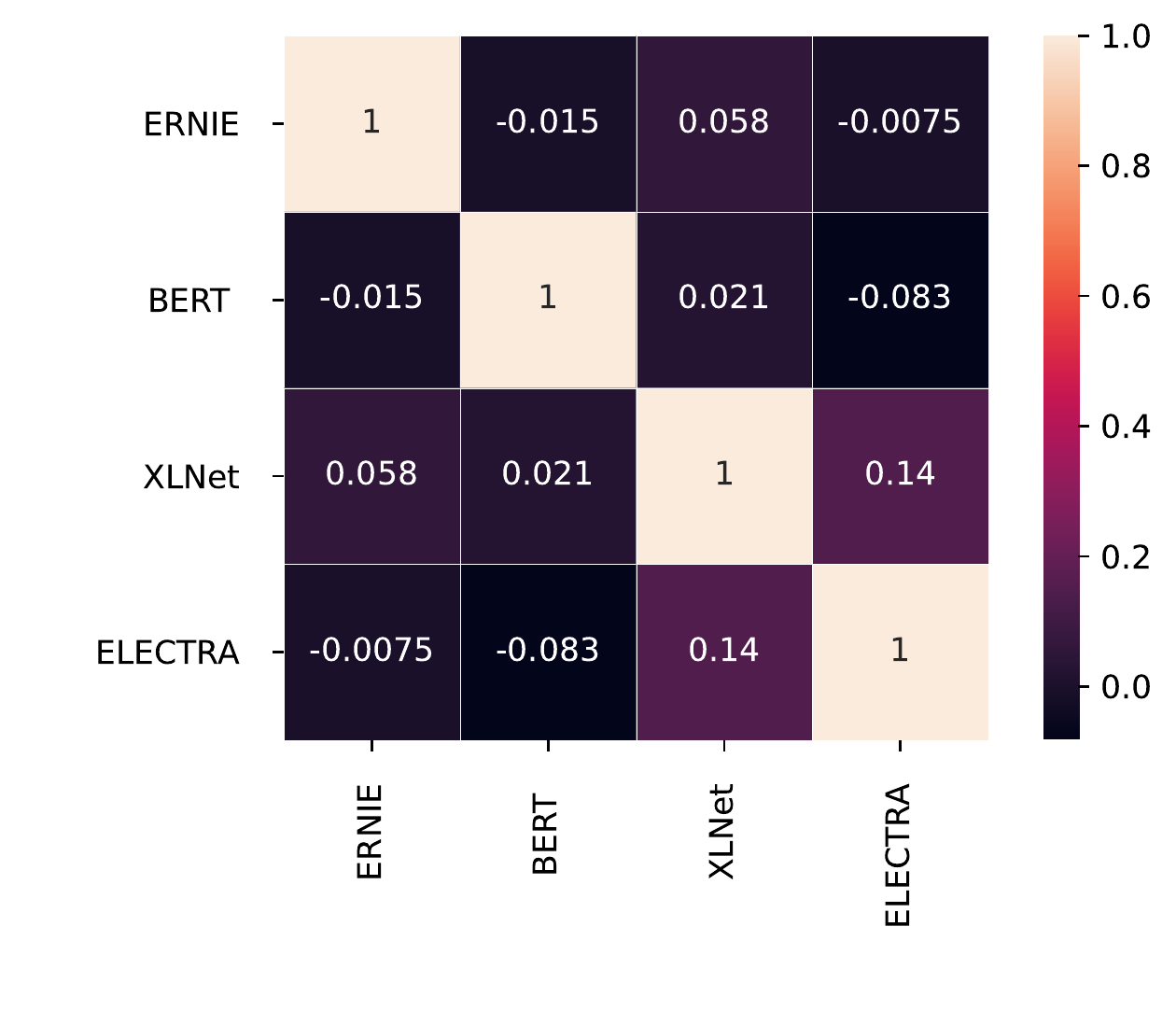} 
\caption{Word-level Gender Bias Comparison of Career Words.}
    
\label{fig.LMCareer}
\end{figure}

\begin{figure}[bt]
\centering
\includegraphics[width=0.9\linewidth]{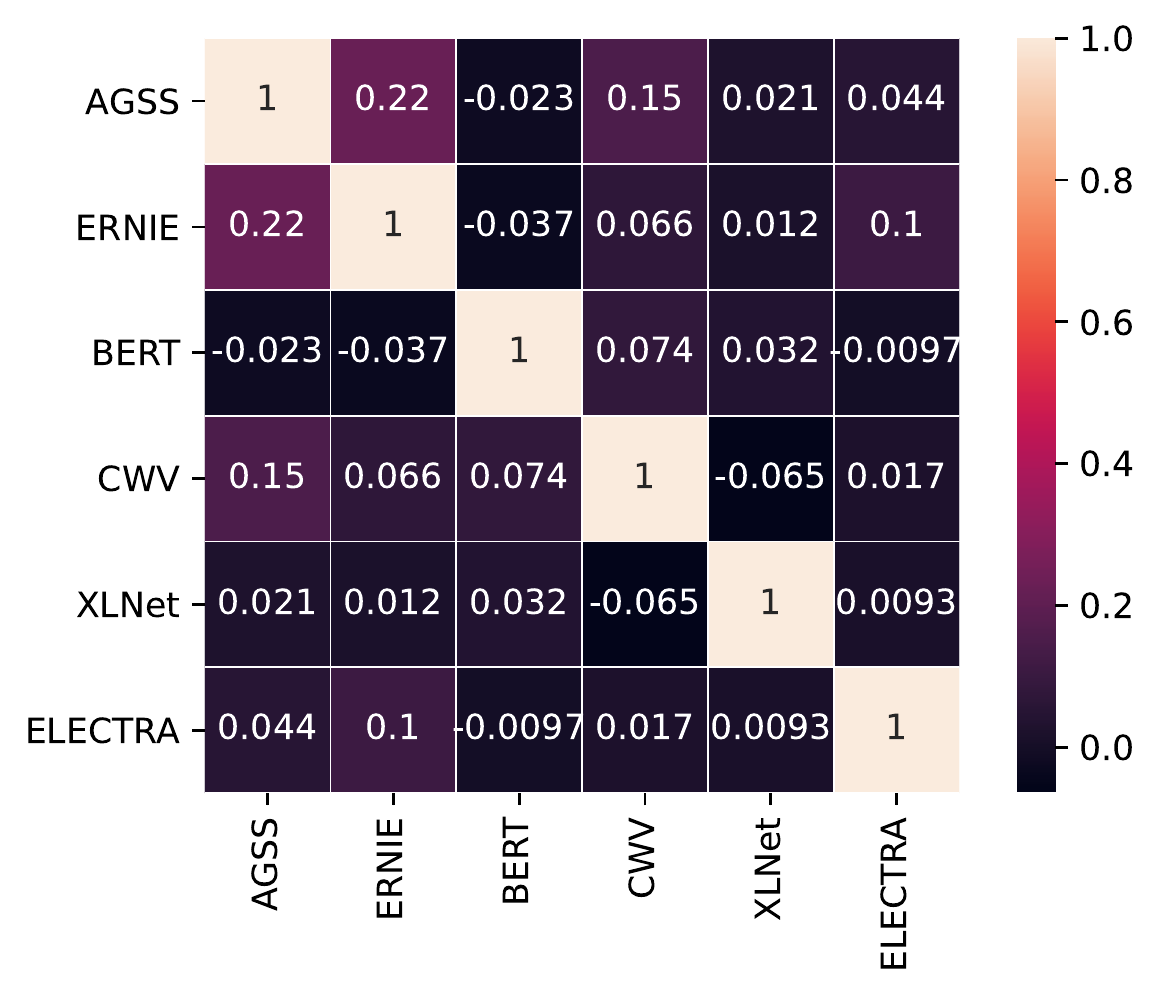} 
\caption{Word-level Gender Bias Comparison of Adjectives. \textbf{CWV} denotes the Chinese Word Vectors trained using mixed-large corpus proposed by \citeauthor{qiu2018revisiting}.}
    
\label{fig.LMAdjective}
\end{figure}

\begin{figure}[bt]
\centering

\includegraphics[width=0.9\linewidth]{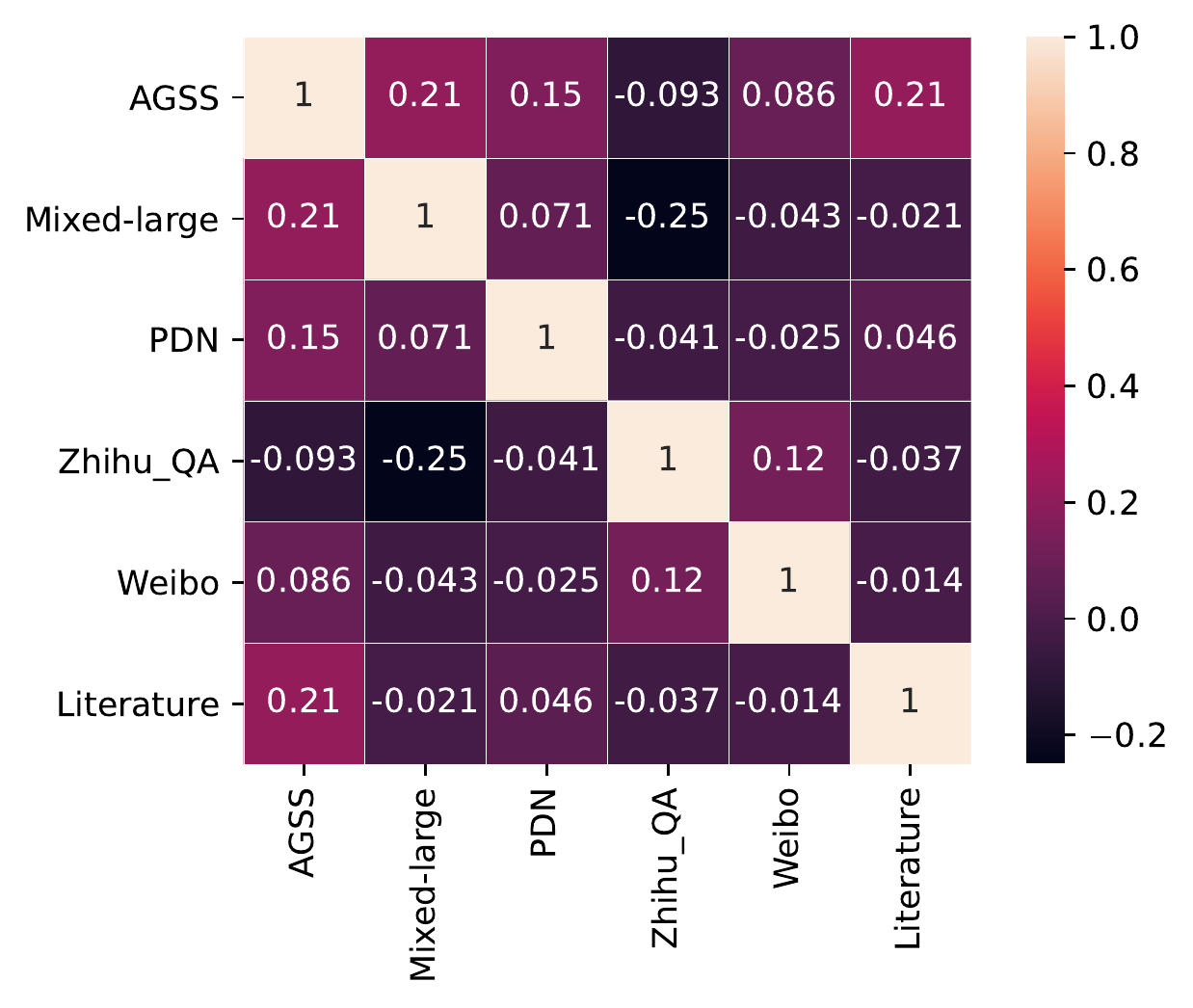} 
\caption{Word-level Gender Bias Comparison of Adjectives of Language Models Pre-trained by Different Corpus. \textbf{PDN} denotes the People's Daily News Corpus.}
    
\label{fig.LMCorpus}
\end{figure}

\begin{figure*}[bt]

\centering

\begin{subfigure}{0.23\linewidth}
\includegraphics[width=\linewidth]{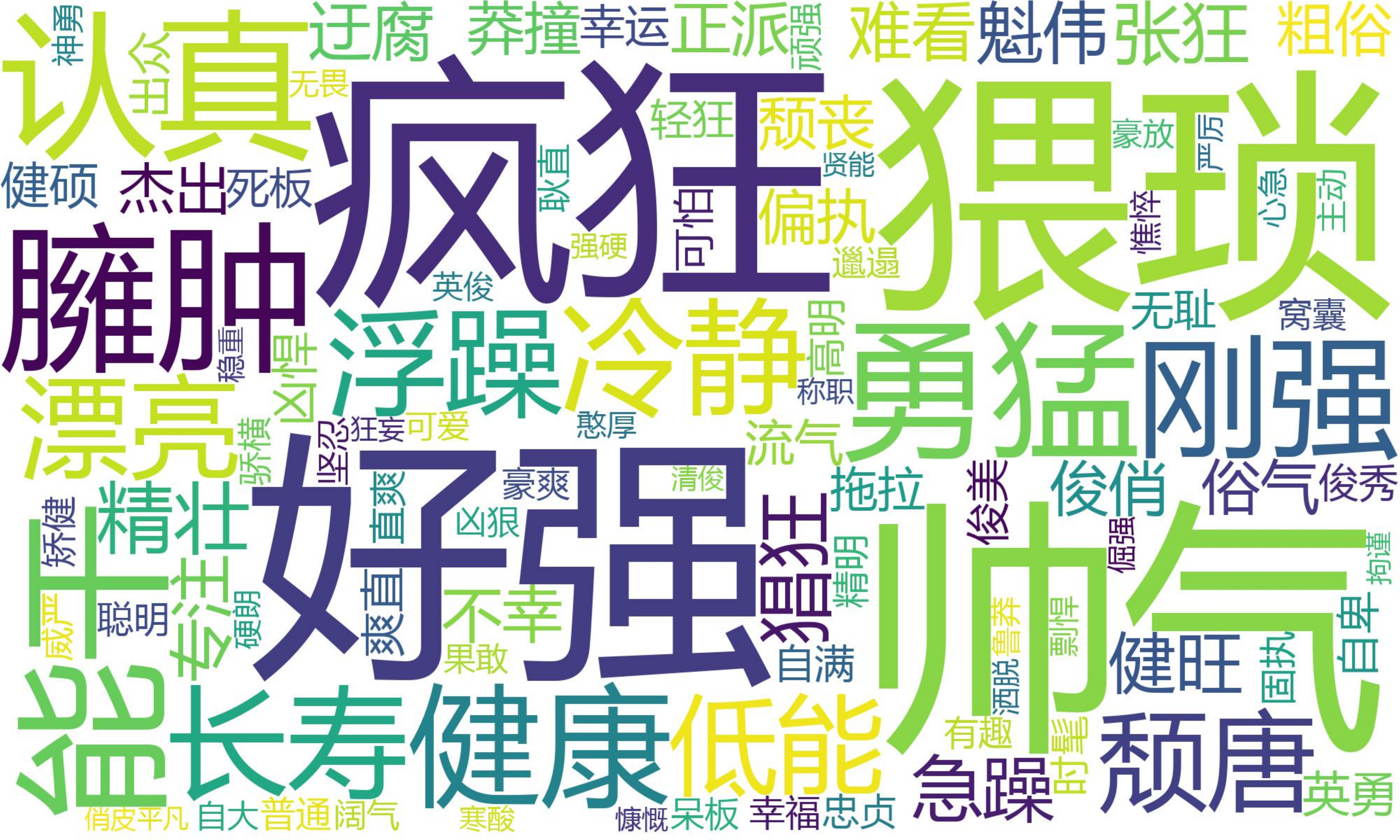} 
\subcaption{Ch-Ernie-Man-Adj}\label{fig:wc_a}
\end{subfigure}
\begin{subfigure}{0.23\linewidth}
\includegraphics[width=\linewidth]{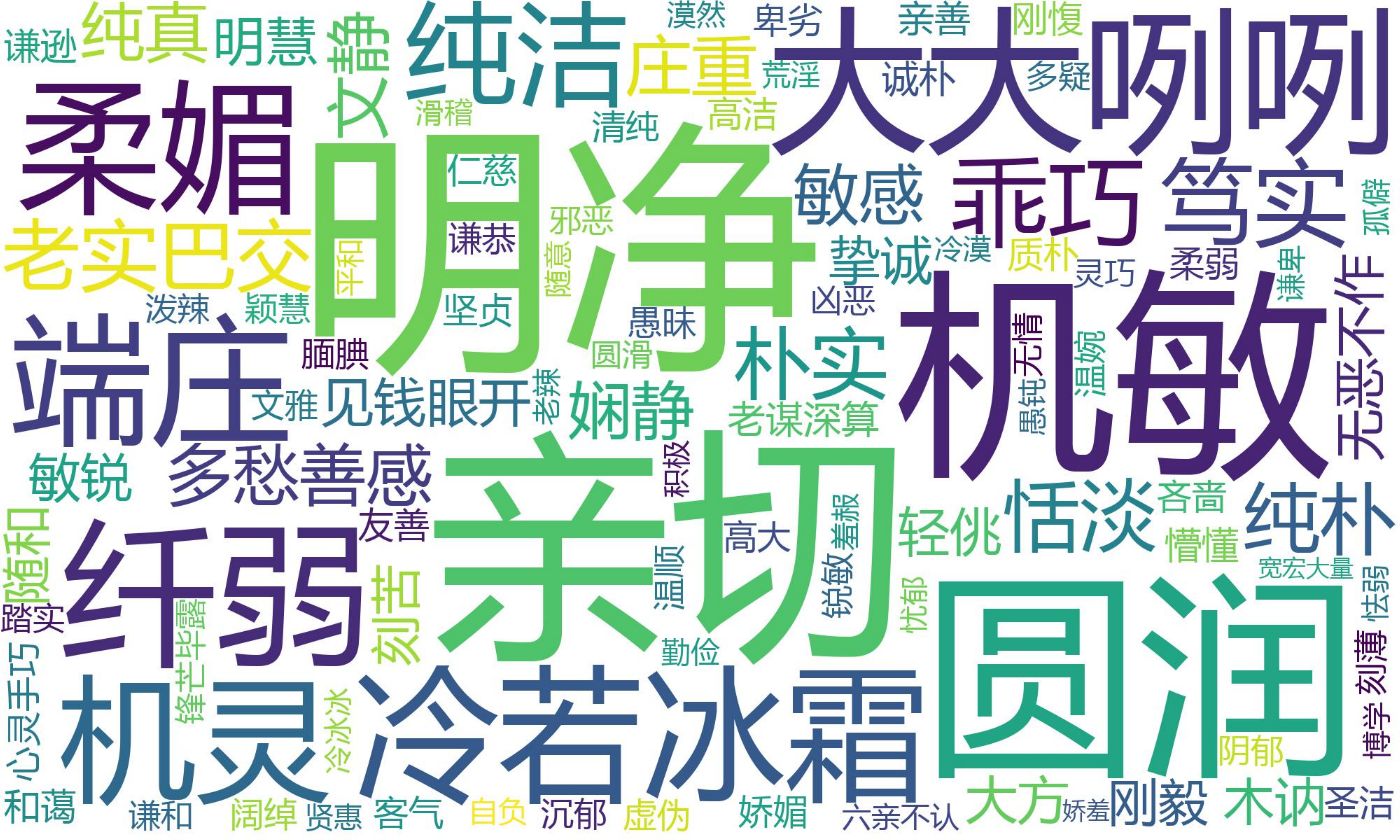} 
\subcaption{Ch-Ernie-Woman-Adj}\label{fig:wc_b}
\end{subfigure}
\begin{subfigure}{0.23\linewidth}
\includegraphics[width=\linewidth]{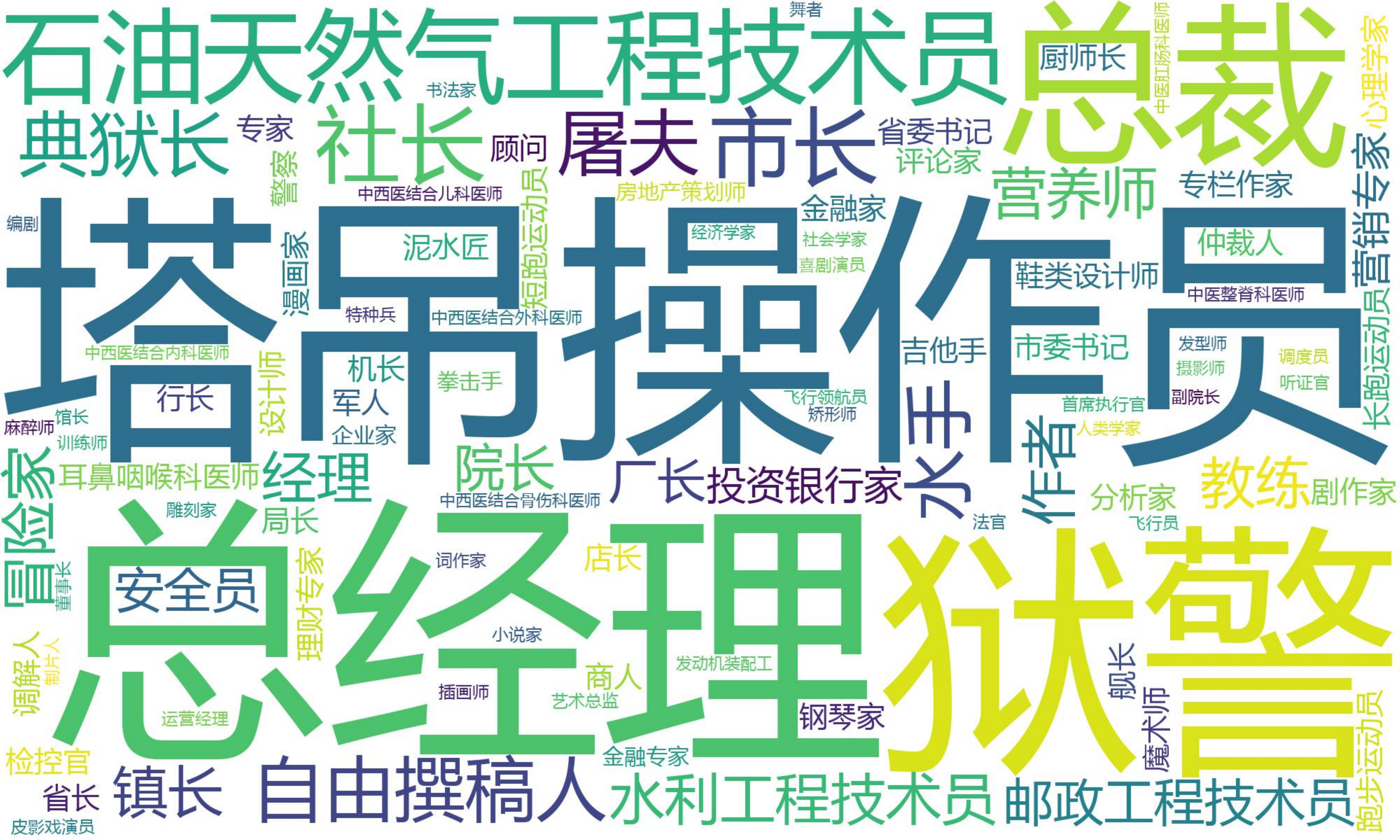}
\subcaption{Ch-Ernie-Man-Career}\label{fig:wc_c}
\end{subfigure}
\begin{subfigure}{0.23\linewidth}
\includegraphics[width=\linewidth]{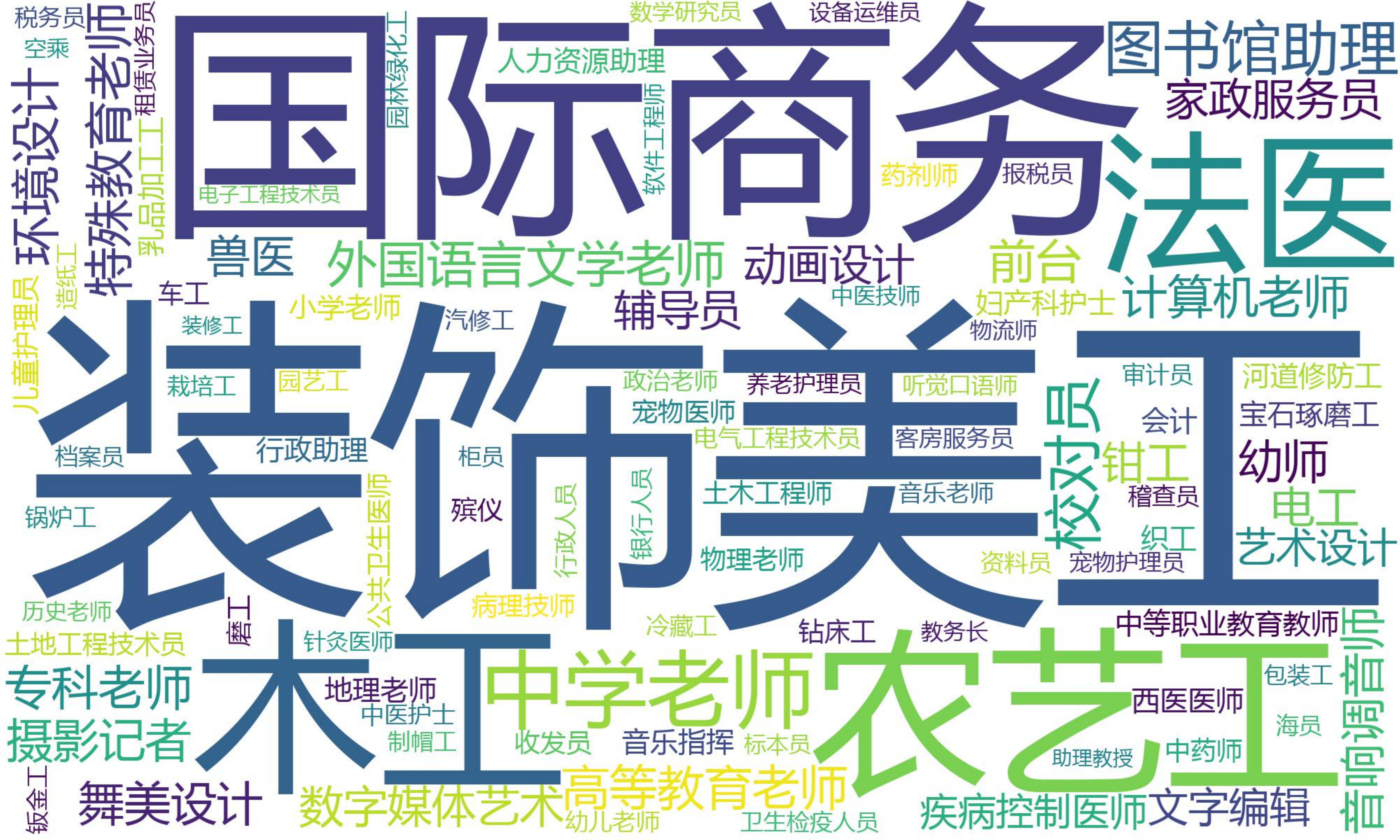} 
\subcaption{Ch-Ernie-Woman-Career}\label{fig:wc_d}
\end{subfigure}
\\ \vspace{1mm}
\begin{subfigure}{0.23\linewidth}
\includegraphics[width=\linewidth]{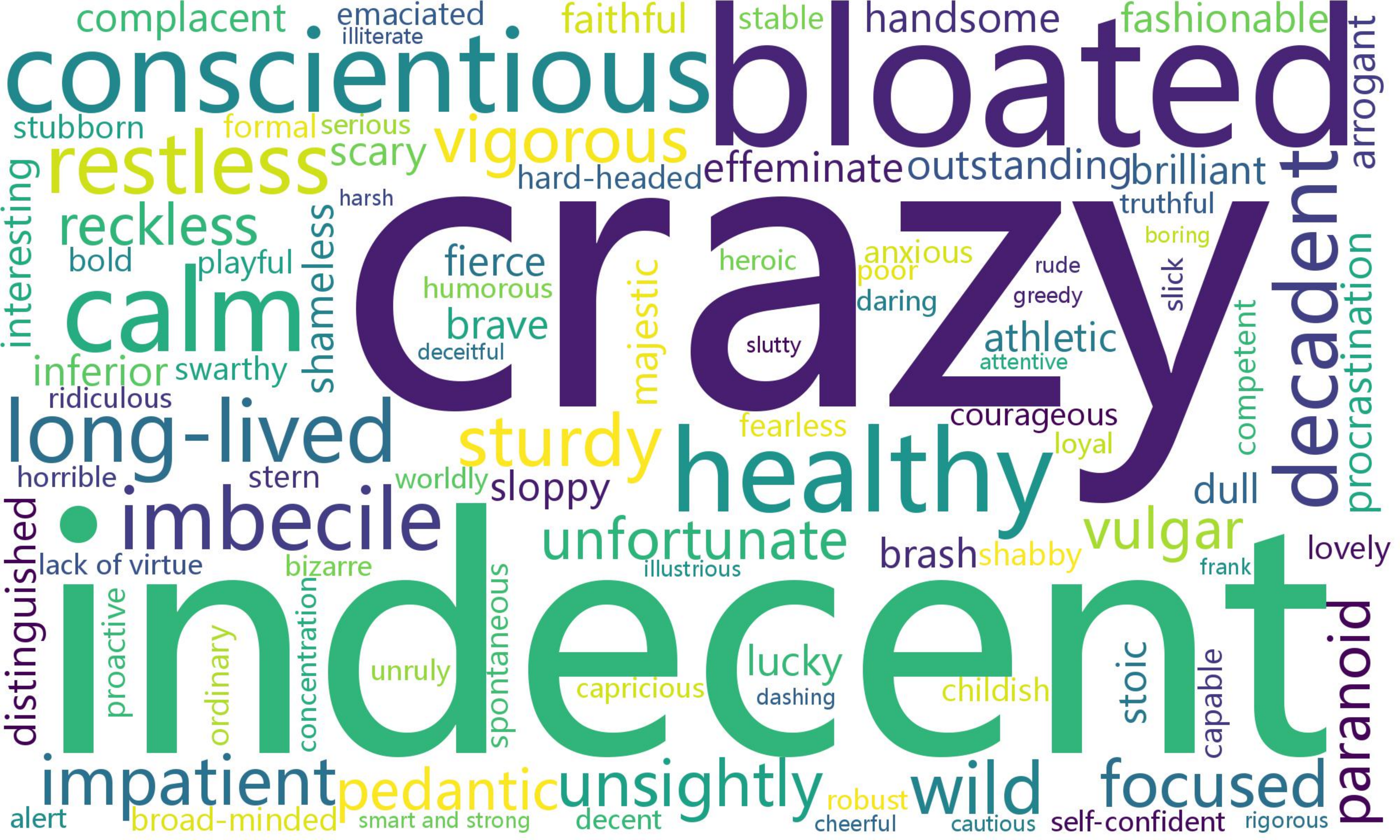} 
\caption{En-Ernie-Man-Adj}\label{fig:wc_e}
\end{subfigure}
\begin{subfigure}{0.23\linewidth}
\includegraphics[width=\linewidth]{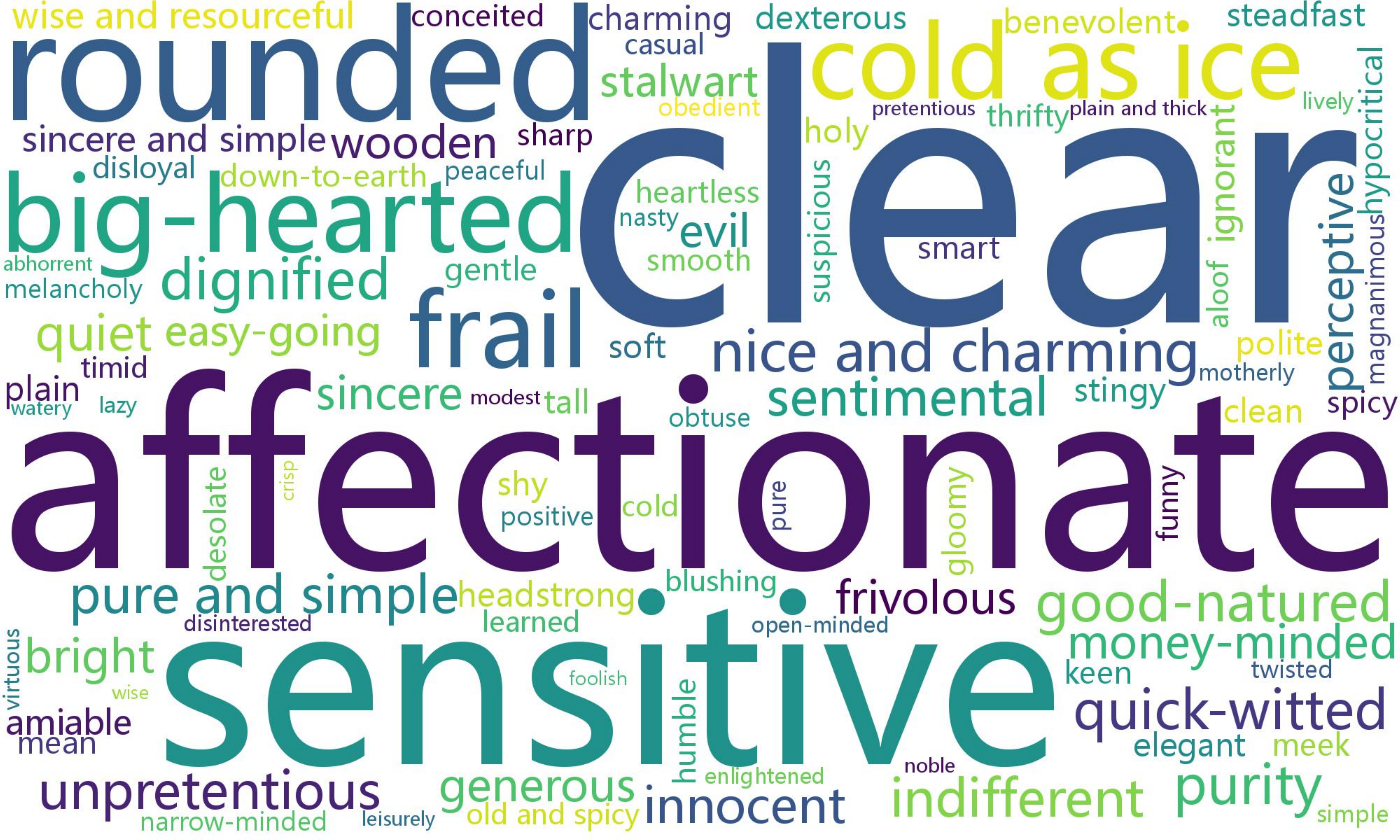} 
\caption{En-Ernie-Woman-Adj}\label{fig:wc_f}
\end{subfigure}
\begin{subfigure}{0.23\linewidth}
\includegraphics[width=\linewidth]{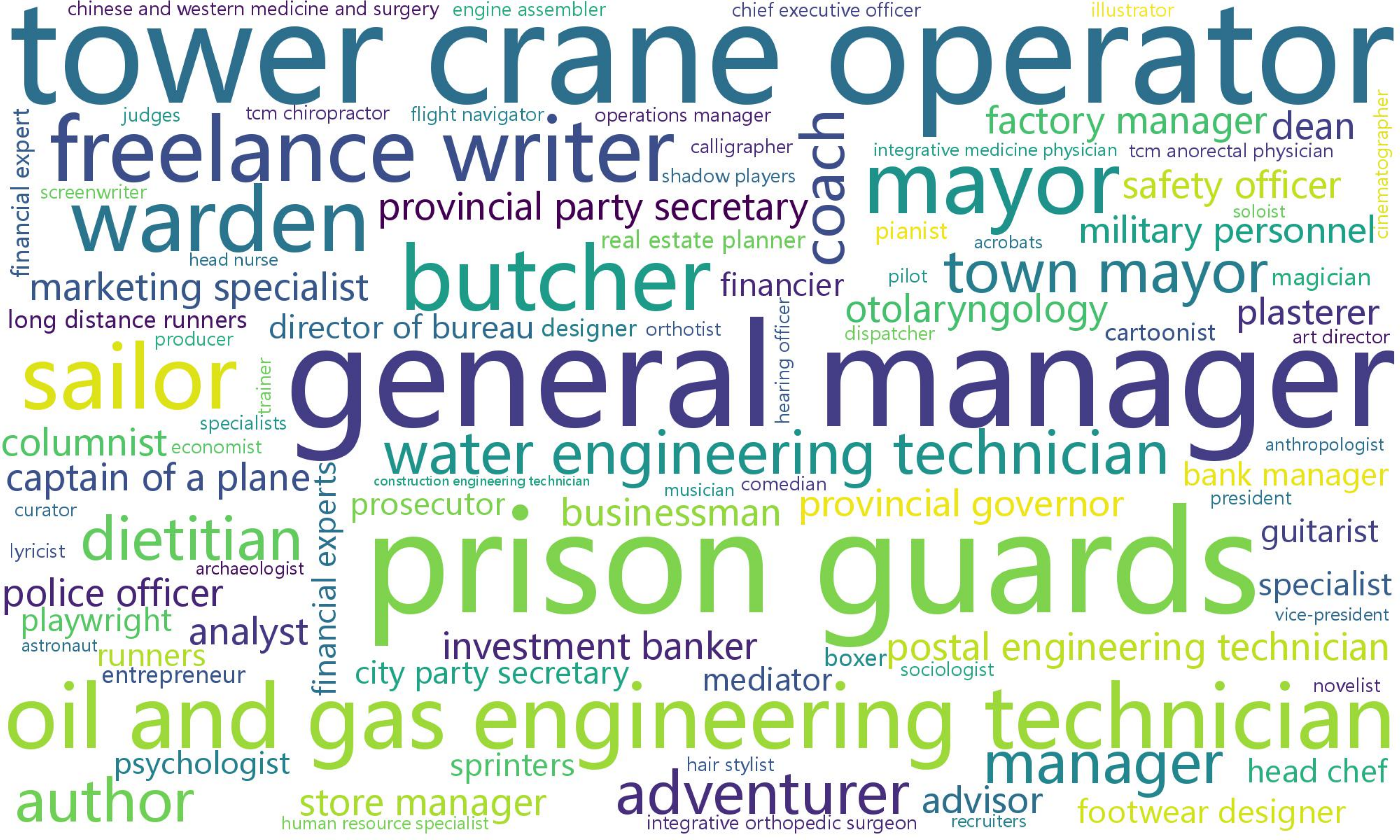} 
\caption{En-Ernie-Man-Career}\label{fig:wc_g}
\end{subfigure}
\begin{subfigure}{0.23\linewidth}
\includegraphics[width=\linewidth]{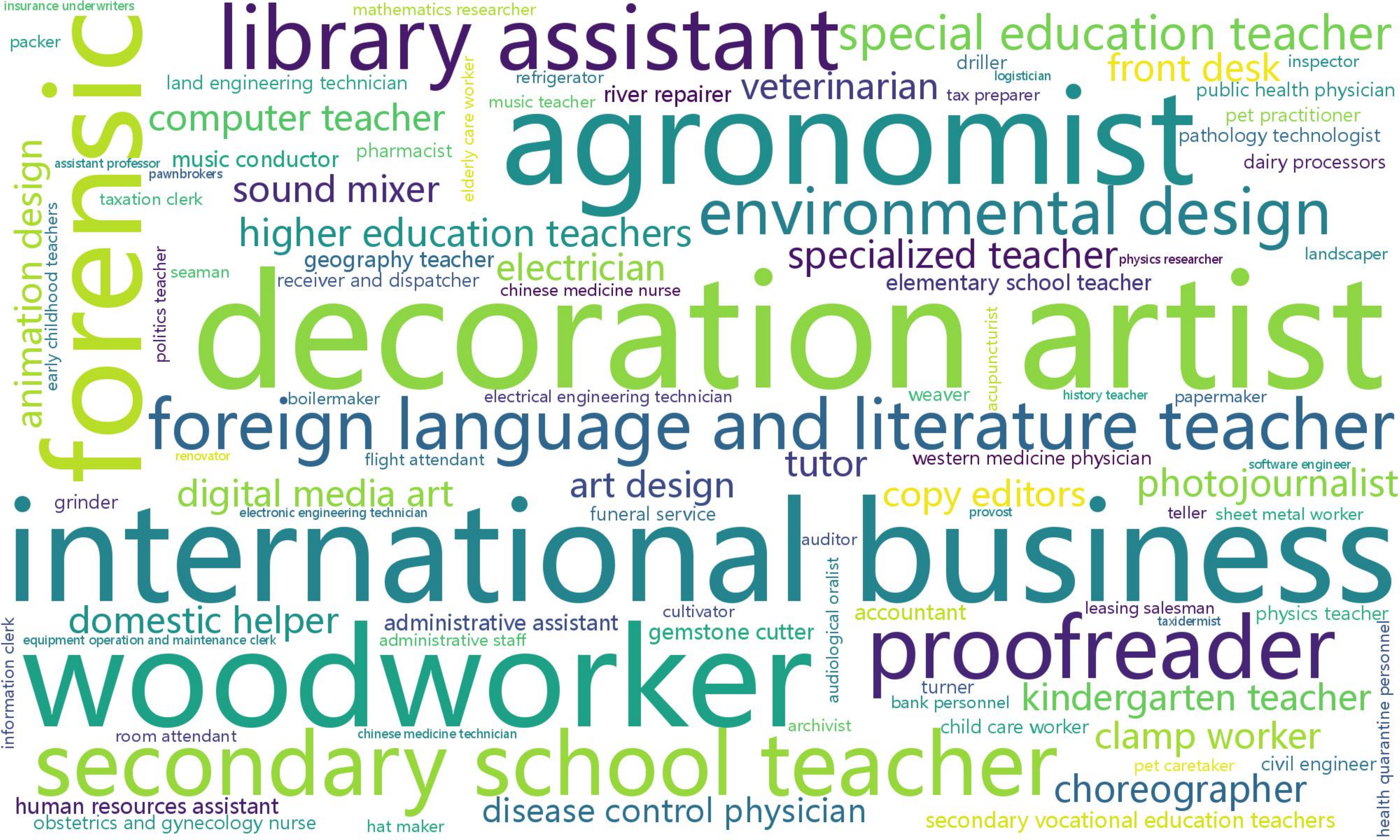} 
\caption{En-Ernie-Woman-Career}\label{fig:wc_h}
\end{subfigure}
\\ \vspace{1mm}
\begin{subfigure}{0.23\linewidth}
\includegraphics[width=\linewidth]{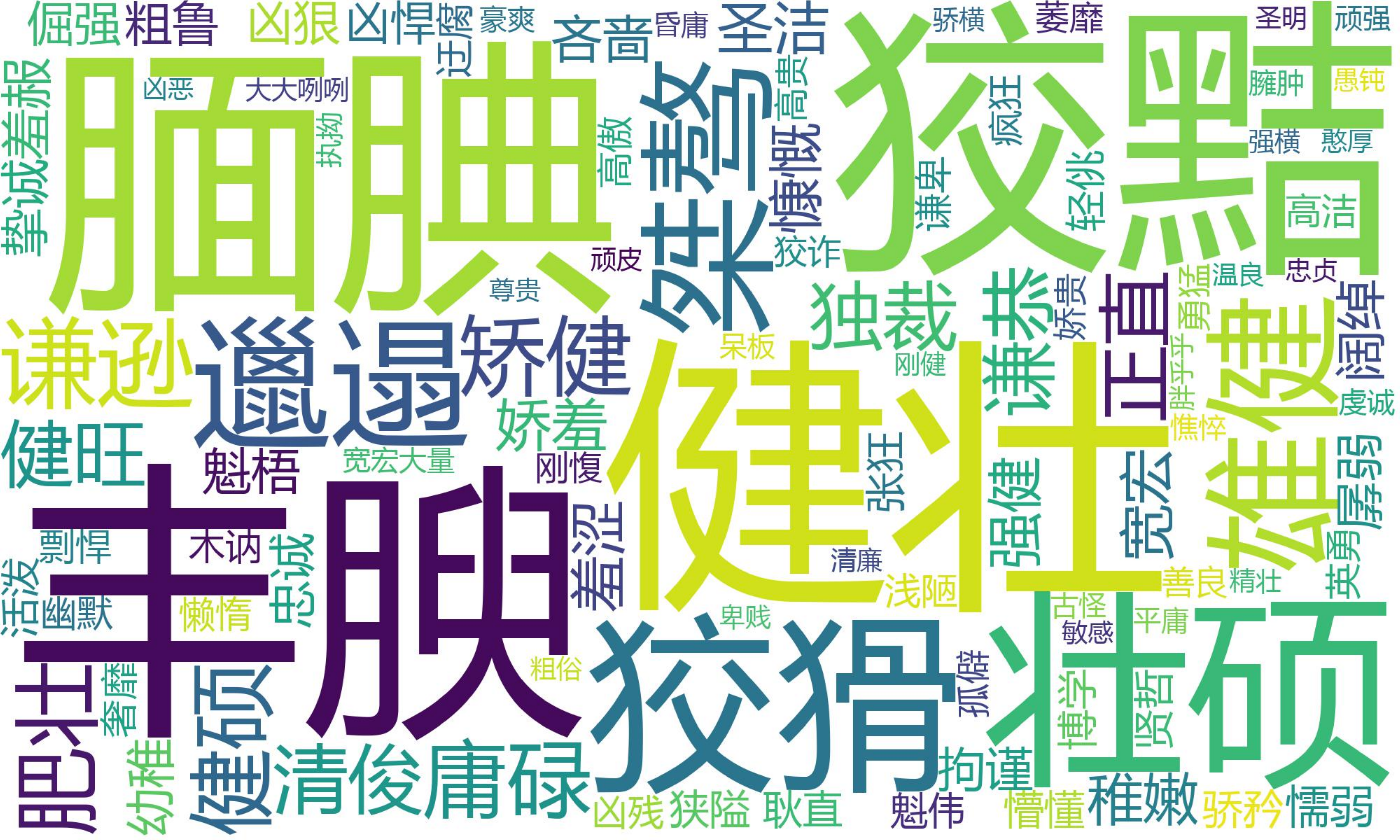} 
\caption{Ch-XLNet-Man-Adj}\label{fig:wc_a}
\end{subfigure}
\begin{subfigure}{0.23\linewidth}
\includegraphics[width=\linewidth]{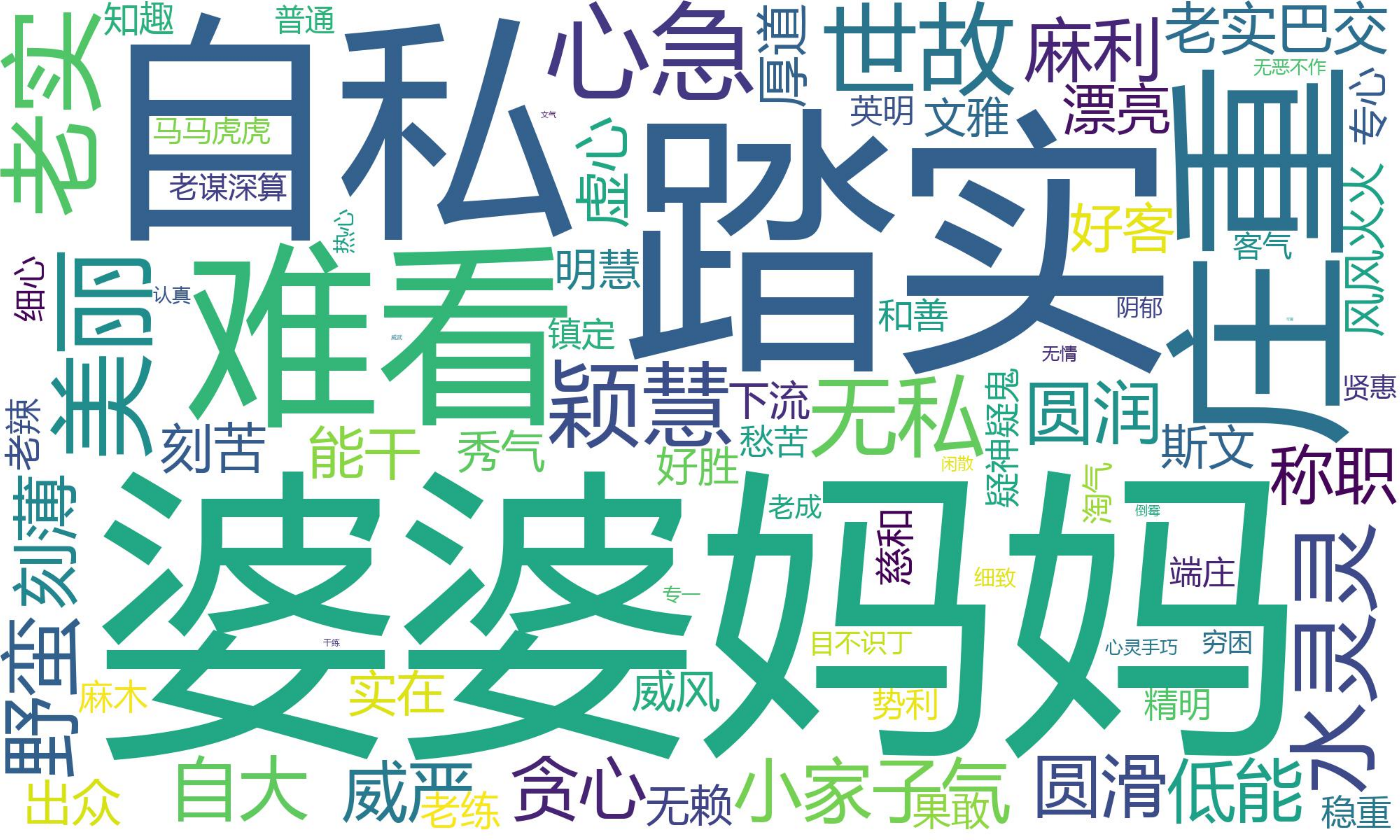} 
\caption{Ch-XLNet-Woman-Adj}\label{fig:wc_b}
\end{subfigure}
\begin{subfigure}{0.23\linewidth}
\includegraphics[width=\linewidth]{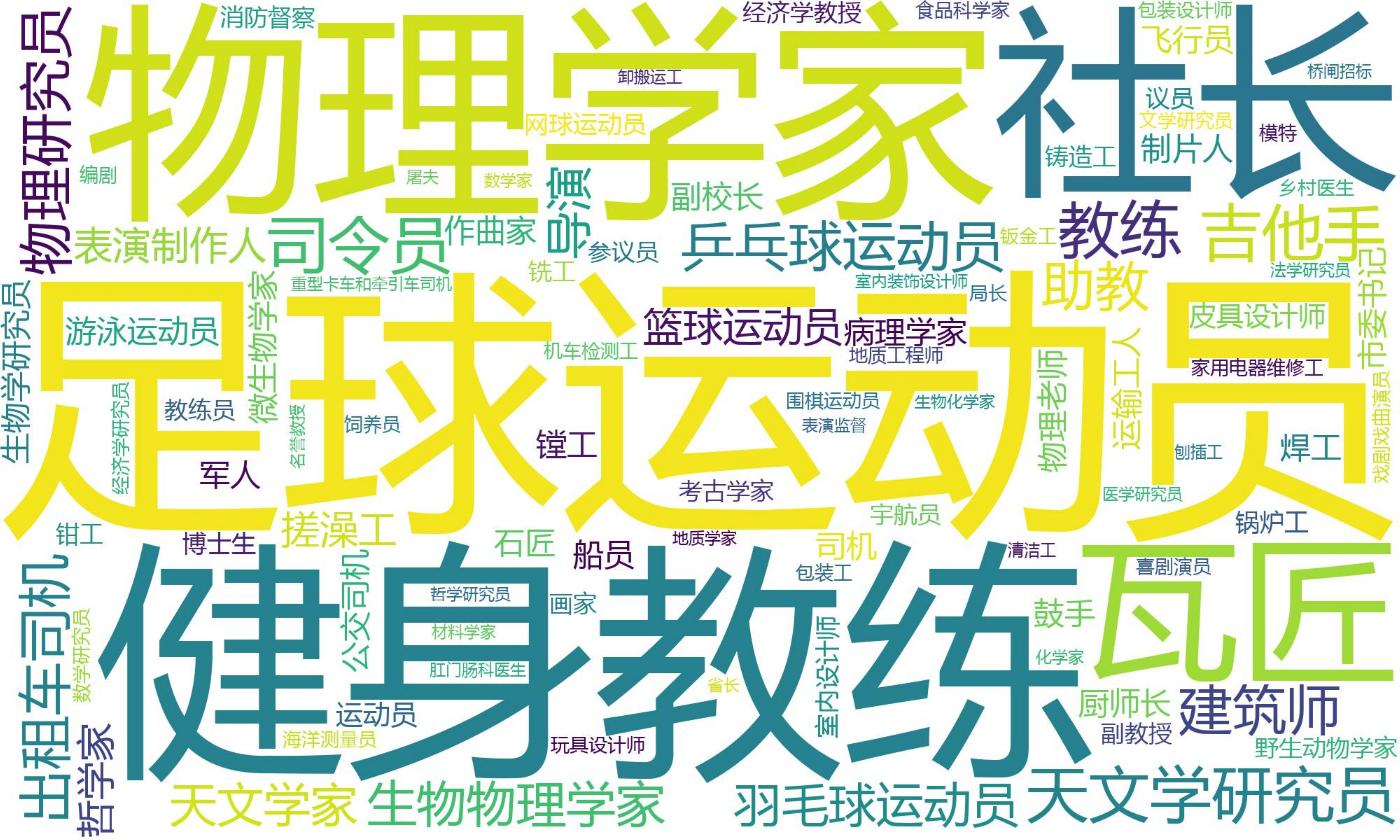}
\caption{Ch-XLNet-Man-Career}\label{fig:wc_c}
\end{subfigure}
\begin{subfigure}{0.23\linewidth}
\includegraphics[width=\linewidth]{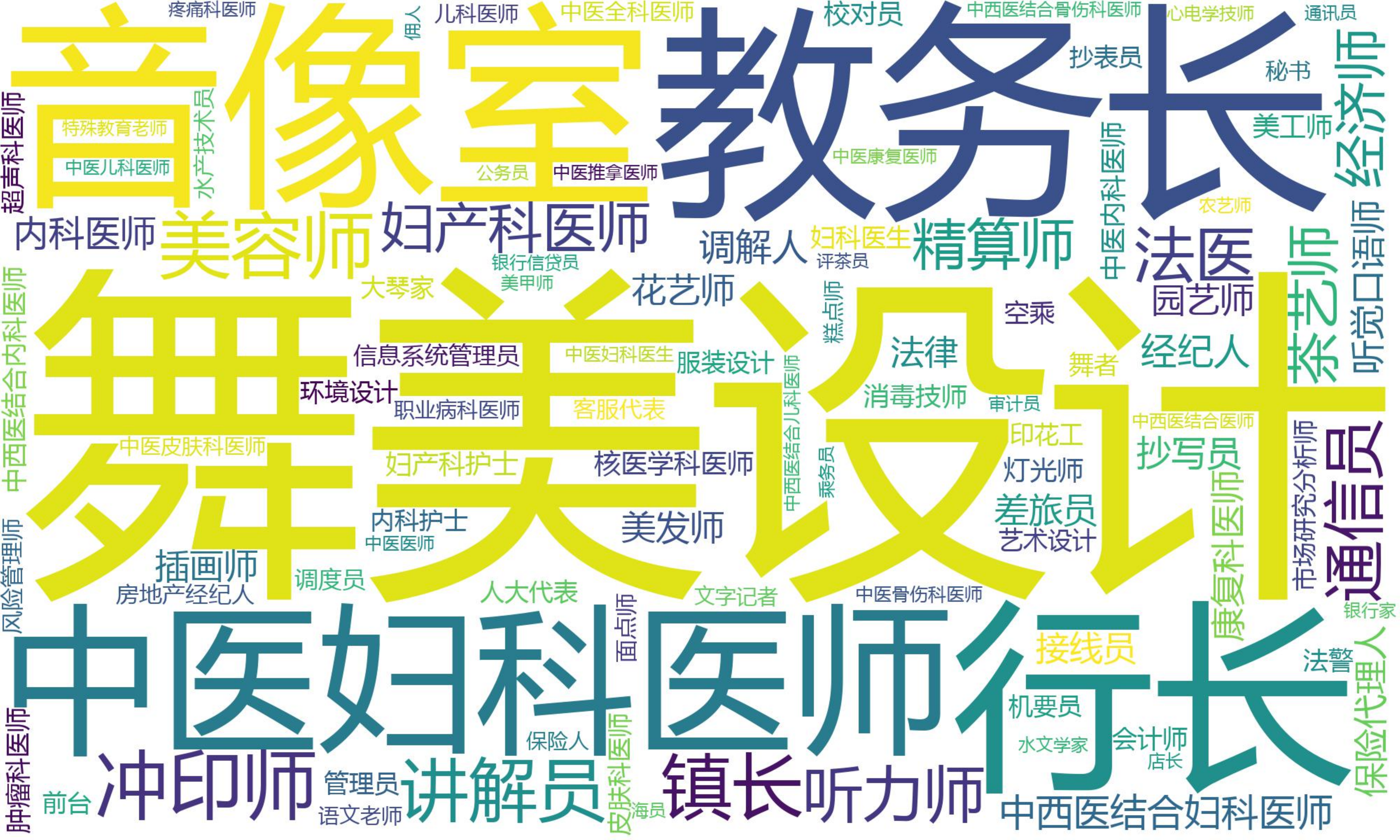} 
\caption{Ch-XLNet-Woman-Career}\label{fig:wc_d}
\end{subfigure}
\\ \vspace{1mm}
\begin{subfigure}{0.23\linewidth}
\includegraphics[width=\linewidth]{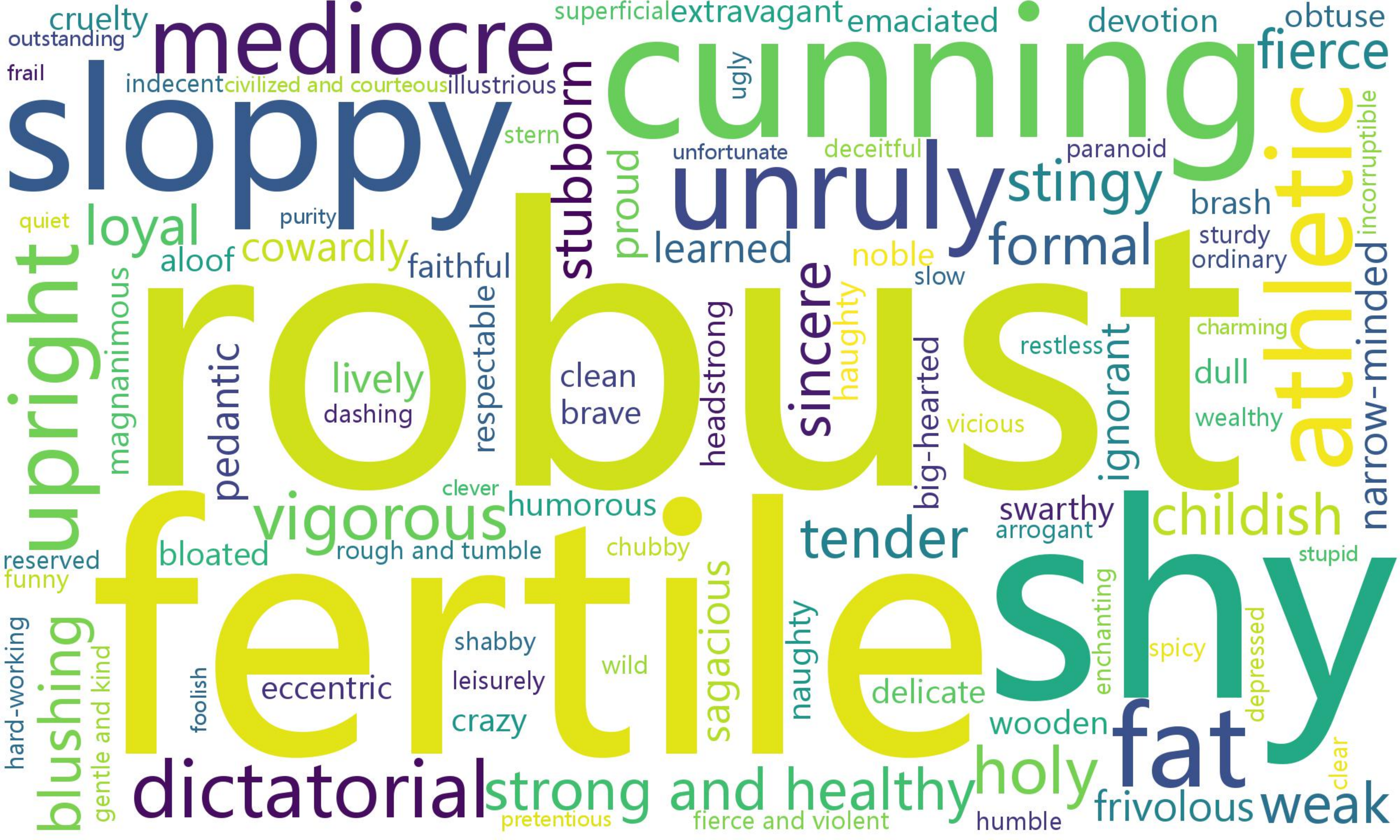} 
\caption{En-XLNet-Man-Adj}\label{fig:wc_e}
\end{subfigure}
\begin{subfigure}{0.23\linewidth}
\includegraphics[width=\linewidth]{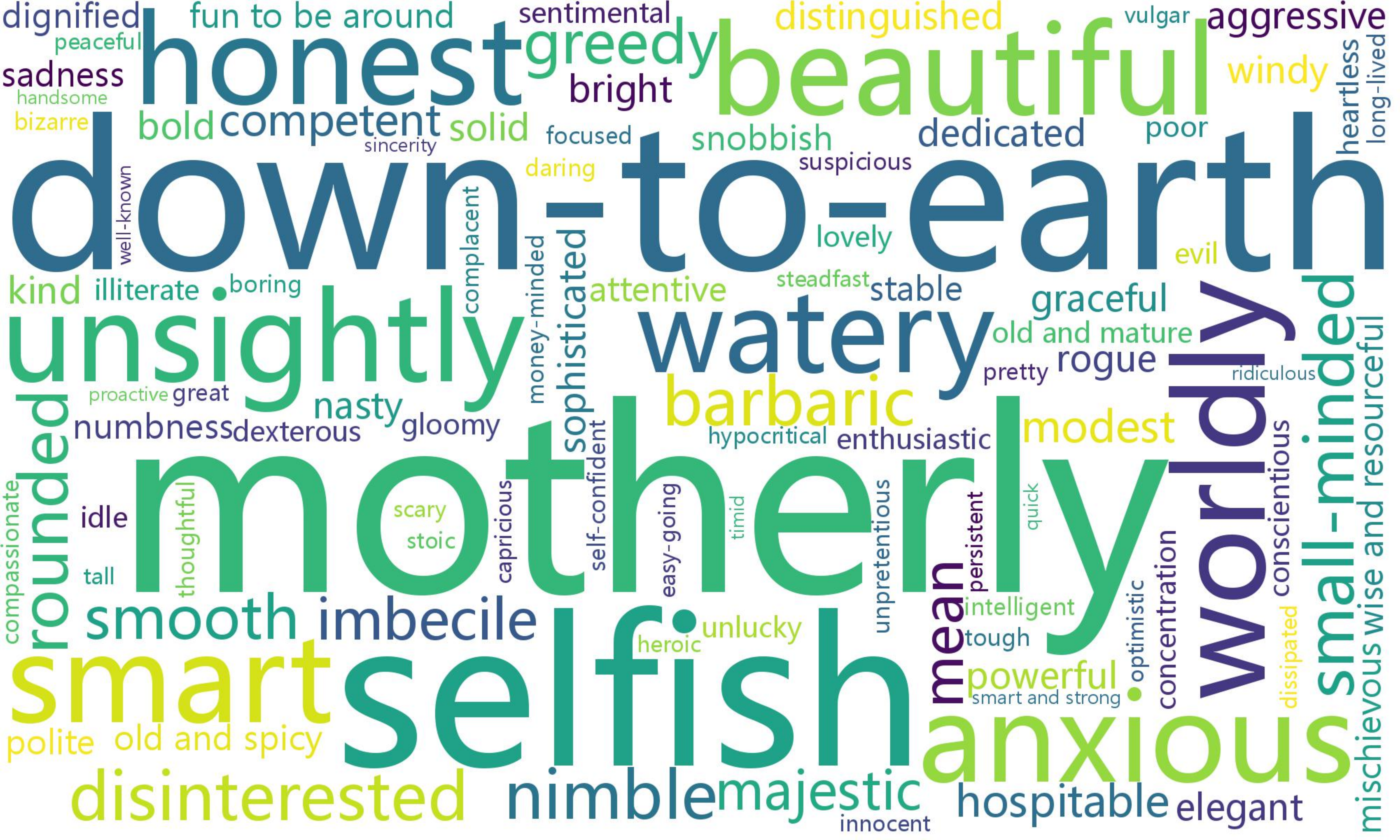} 
\caption{En-XLNet-Woman-Adj}\label{fig:wc_f}
\end{subfigure}
\begin{subfigure}{0.23\linewidth}
\includegraphics[width=\linewidth]{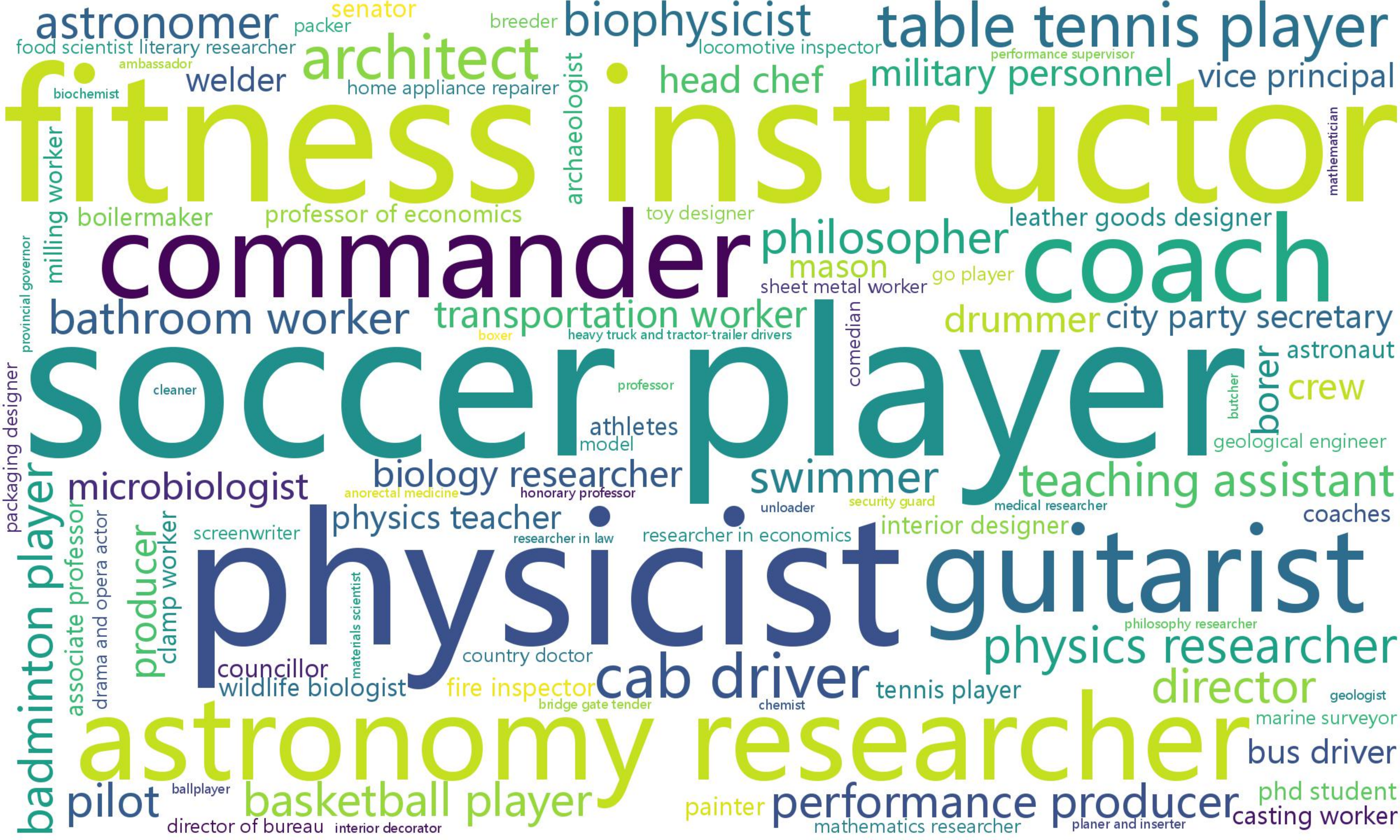} 
\caption{En-XLNet-Man-Career}\label{fig:wc_g}
\end{subfigure}
\begin{subfigure}{0.23\linewidth}
\includegraphics[width=\linewidth]{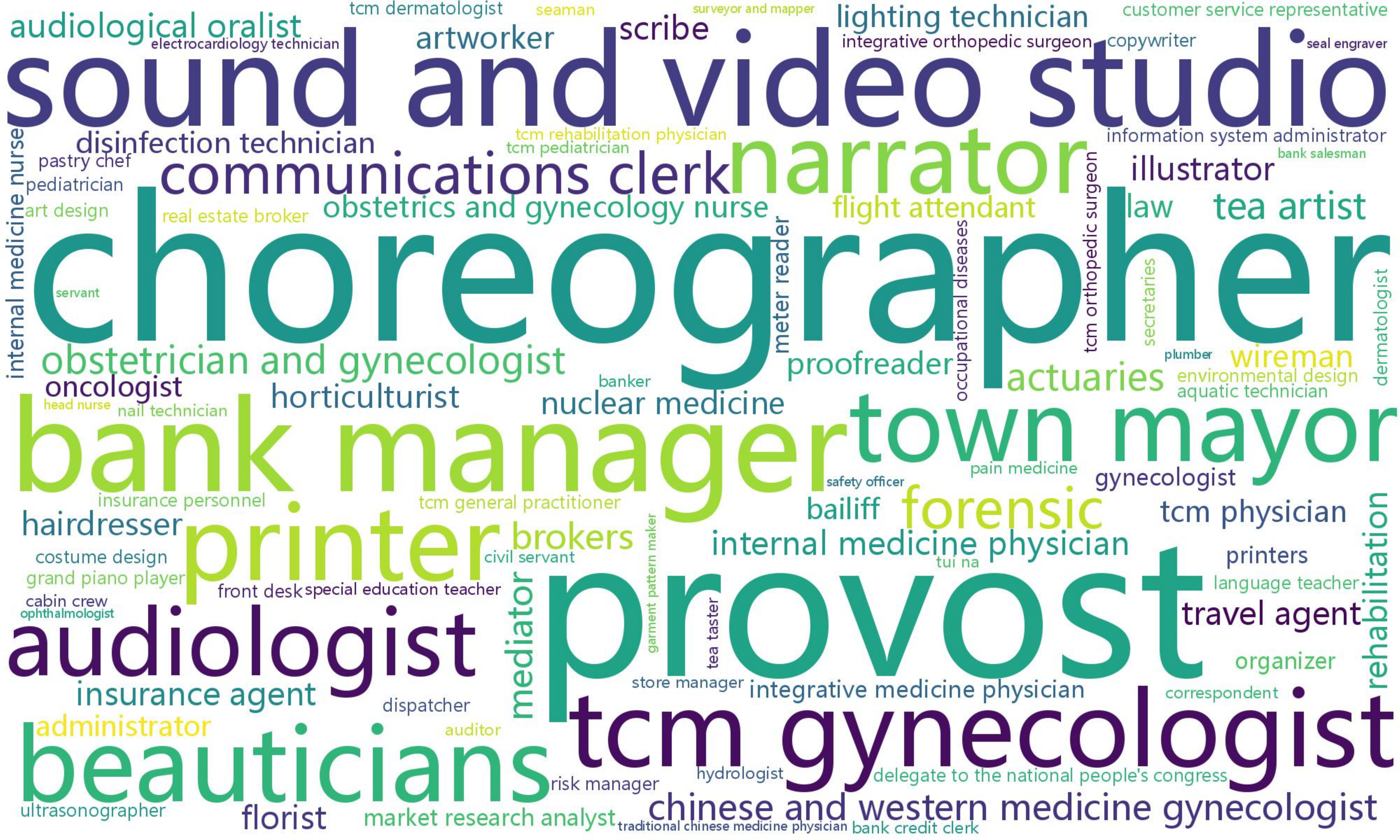} 
\caption{En-XLNet-Woman-Career}\label{fig:wc_h}
\end{subfigure}
\\
\vspace{1mm}
\caption{Example Word Cloud Analysis of Ernie and Chinese-XLNet. \textbf{Ch} denotes Chinese. \textbf{En} denotes words' English translation. \textbf{Man} and \textbf{Woman} separately denote words with embedding closer to man and woman. \textbf{Adj} denotes adjectives. \textbf{Career} denotes career words.}
\label{fig:wc}
\end{figure*}

We conduct described comparison of adjectives between AGSS as a golden standard \cite{zhu2020great}, Ernie \cite{zhang2019ernie}, Chinese Word Vectors trained by mixed corpus \cite{qiu2018revisiting}, and Chinese-XLNet, Chinese-Bert, and Chinese-Electra proposed te{cui-etal-2020-revisiting} to produce Fig.~\ref{fig.LMAdjective}.
We conduct described comparison of career words between Ernie \cite{zhang2019ernie}, and Chinese-XLNet, Chinese-Bert, and Chinese-Electra proposed te{cui-etal-2020-revisiting} to produce Fig.~\ref{fig.LMCareer}.
The described experiments on career words is not conducted with the Chinese Word Vectors trained by mixed corpus, because an observing number of career words are missing in its dictionary.

We don't provide a golden standard vector \cite{srivastava2022beyond} since they didn't provide a manual gender bias analysis about the career words.
We also conduct described comparison on adjectives in Chinese Word Vectors pre-trained by different corpus, including Mixed-large corpus, People's Daily News, Zhihu QA dataset, Weibo, and Chinese literature dataset to produce Fig.~\ref{fig.LMCorpus} and analyze the learned gender bias difference caused by using different datasets for pretraining the language model.

\subsection{Discussion}

There exists observing gender bias in the open-source Chinese language models, especially in Ernie and Chinese Word Vectors according to Fig.~\ref{fig.LMAdjective}. 
We hypothesize that the observation is highly related to the corpus used. 
\citeauthor{cui2020revisiting} claim that their used corpus is a combination of ChineseWiki, and some other universal Chinese datasets, including encyclopedia, news, and QA dataset. In sharp contrast, Ernie and Chinese Word Vectors use corpus, which contains sentences from literature, forum, and other social media, which may lead to a gender-biased model.

According to Fig.~\ref{fig.LMCorpus}, People's Daily News, and Chinese literature corpora contain observing gender bias. The observation indicates that researchers should be more careful about using literature data while training a language model. We also hypothesize that this is caused by the literature corpus and People's Daily News, which contains more descriptive expressions.

\section{Corpus}

\subsection{Word Cloud Analysis}
\label{sec:appen_cor_cloud}
We provide word cloud analysis of Ernie and Chinese-Electra in the section about adjectives and career words.
More available word cloud analysis will be available in our public repository.
The words are ranked according to the absolute value of their gender bias score calculated along the method used by \citeauthor{Bolukbasi2016ManIT,jiao2021gender}. 
There is a noticeable word-level gender stereotype according to the word cloud. 
For example, a man is robust and a woman is motherly, a man is suitable for a fitness instructor and a woman is suitable for a choreographer.
We also conduct word cloud analysis for language models pre-trained by different corpora.

\subsection{Quality Monitoring and Control}\label{sec:appen_qm}
We used a standardized operating method and educated our annotators to achieve high-quality annotations as follows:

(1). \textbf{Annotators} \quad We have 6 annotators, which were all native speakers of Chinese.
Annotators were only qualified to do the annotation if they went through several societal \cite{king2021gender, xu2019cinderella} and computer science research works \cite{sun2019mitigating, zhao2018learning} about gender bias before the annotation procedure. 
All annotators held a bachelor's degree. 
\citeauthor{waseem-2016-racist} points out that expert annotators are more cautious and can improve the corpus quality with a large margin, which proves the necessity of our training procedure. 
We also kept the number of male and female annotators equal.

(2). \textbf{Gender Equality of Raw Corpus} \quad In the raw data collection procedure, we keep the number of man-related keywords and woman-related keywords equal and make the number of samples recalled according to different keywords balanced. As a result, the raw data and the final data should hold gender equality.

(3). \textbf{Annotation Procedure} \quad Our annotation procedure is separated into two stages. 
In the first stage, annotators are encouraged to not enter any samples that they are not certain about. 
In the second stage, we have annotators cross-checking annotations.
We did not enter any contradictory samples.

(4). \textbf{Inter-annotator Agreement} \quad Given the domain and purpose of the dataset, we want to build the dataset as high quality as possible. After an initial annotation round with 6 annotators, we also report inter-annotator agreement in Table \ref{iaa}. to verify annotation reliability, where the IAA among three annotators on bias classification, detection, and mitigation is 0.802, 0.935, and 0.987, respectively.

\begin{table}[h!]
    
    \centering
   \resizebox{\linewidth}{!}{ 
    \begin{tabular}{c|ccc}
    \hline
    & \textbf{Classification} & \textbf{Detection} & \textbf{Mitigation}\\
    \cline{1-4}
    IAA & 0.802& 0.935 & 0.987\\
    \hline
    \end{tabular}}
    \caption{Inter-Annotator Agreement (IAA)}
    \label{iaa}
    
\end{table}

\begin{figure*}[ht]
\vspace{-10mm}
    \centering
    \includegraphics[width=0.65\linewidth]{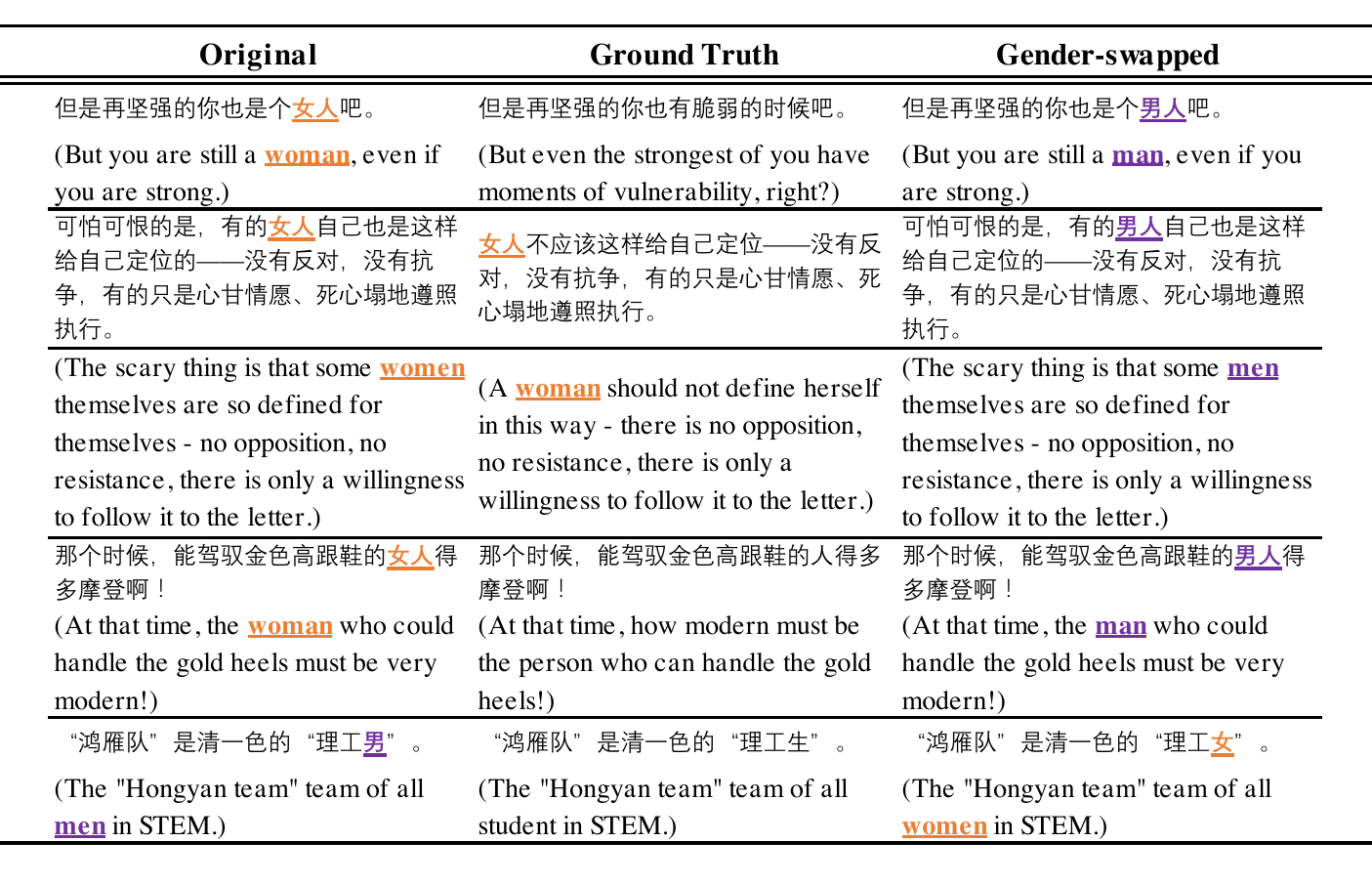}
    \caption{Case Study of Nonsensical Sentences Created by Gender-swapped Methods.}
    \label{fig:CaseStudyProblem}
\end{figure*}
\begin{figure*}[!ht]
\vspace{-5mm}
    \centering
    \includegraphics[width=0.6\linewidth]{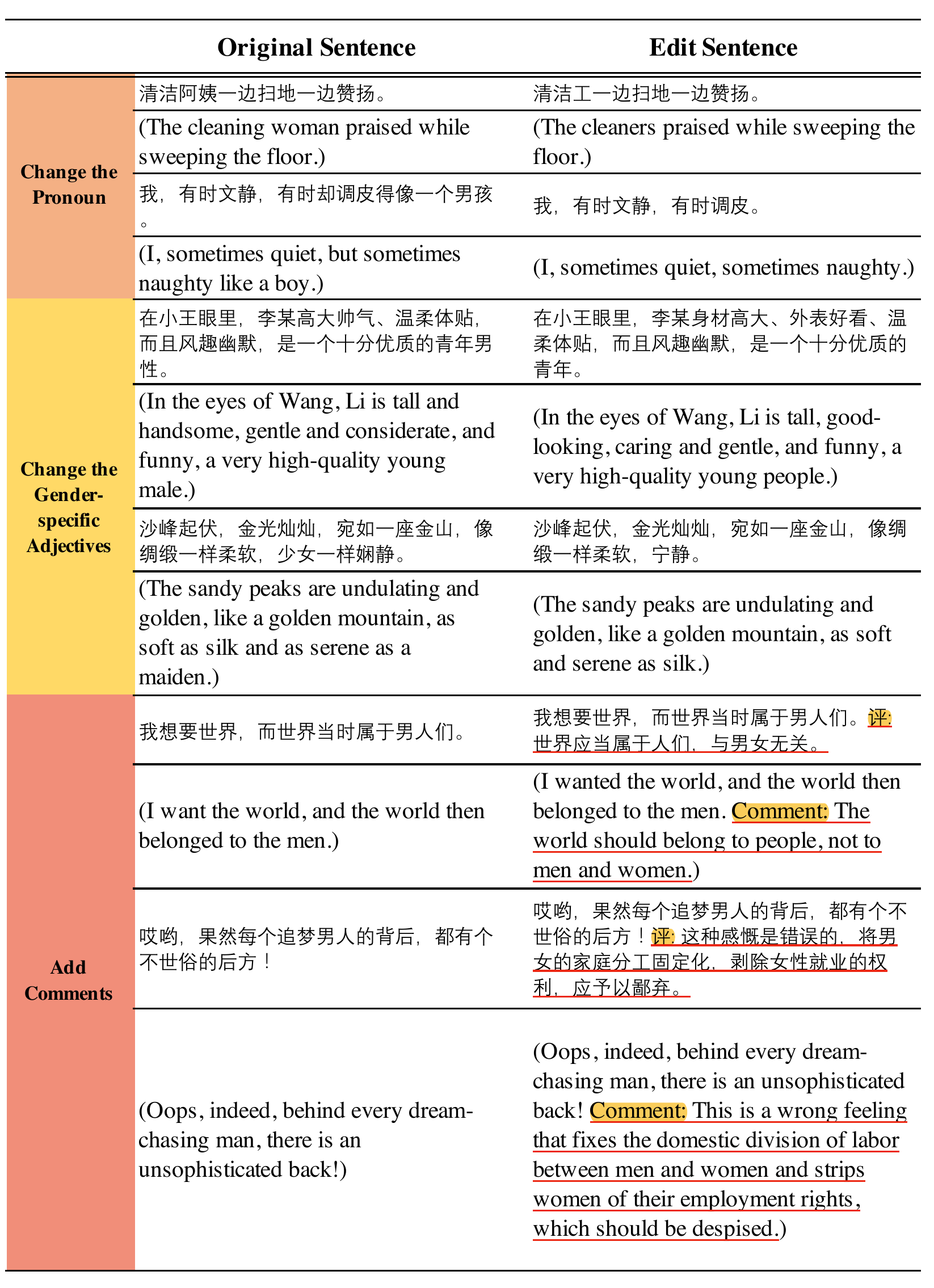}
    \caption{Case Study of Mitigation Annotation Patterns.}
    \label{fig:CaseStudyEditing}
\end{figure*}

\section{Implementation Details}
\label{Appendix.ImplementationDetail}

For \textbf{gender bias classification challenge}, we used finetuned Chinese-BERT-wwm, Chinese-ELECTRA-180g-base, and Chinese-XLNet-base, ~\cite{cui2020revisiting}, and the GPT-3 (Curie) in the in-context paradigm. We first use the train set to save the multiple labeled examples in a document with a specific file ID. Then we use the test sets to perform a classification query on the saved file. The processing time for the classification of gender bias is approximately 1 hour. We calculated the precision, recall, and F1 score to analyze model performance. 

For \textbf{gender bias detection challenge}, we use the same baseline model set as in the classification challenge. We test the performance on both "yes" and "no" detection. 
The detection tasks also use the Classification endpoints of GPT3 (Curie), which requires more time compared to classification as we use a larger dataset for both training and testing.

For \textbf{gender bias mitigation challenge}, we did not provide experiment results of finetuning the largest Davinci (175B) GPT-3 on CORGI-PM 
because of the cost and no observing performance gain comparing Curie and Babbage. 
For finetune experiment setting, we follow the tutorial of GPT-3 official API of the Completion Model and regard the ground truth edits provided by human annotators as the completion of the original sentences.
For the zero-shot experiment setting, we apply GPT-3 editing model and set the instructions as "Eliminate the gender bias contained in the sentence."

For metrics used, on the one hand, we conduct extensive human evaluations from both gender bias and coherence aspects on CORGI-PM. 
For both gender bias and coherence, we shuffled the correction results from human annotators and different models, and asked our annotators to grade the results using the answer range from 1-\textit{not at all} to 7-\textit{extremely gender biased/extremely fluent} without the information of the source.
On the other hand, we provide the automated metrics result\href{https://github.com/Maluuba/nlg-eval}, including BLEU \cite{papineni2002bleu}, ROUGE-L \cite{lin2004rouge}, and METEOR \cite{agarwal2007meteor} of the models on CORGI-PM as well. 
BLEU is the earliest and most widely-used metric for translation and NLG tasks.
METEOR introduces WordNet and other external resources to improve the robustness of the BLEU-based metrics.
ROUGE pays more attention to recall compared to BLEU.

In the gender bias correction task, we fine-tune Ada, Babbage, and Curie models of GPT3 and test the performance using the aforementioned metrics. Fine-tuning on the train set requires 31 minutes on Ada, 35 minutes on Curie, and 43 minutes on Babbage. For the generation process, each fine-tuned model required approximately 30 minutes to complete. For the zero-show paradigm, we use "Eliminate the gender bias in the sentence" as the instruction and use Davinci. Compared with the fine-tuned model doing the sentence correction, the zero-shot paradigm requires more time (approximately 1 hour).

\section{Case Study}
\label{Appendix:CaseStudy}
As shown in Fig.~\ref{fig:CaseStudyProblem}, gender-swapped methods suffer from mitigating gender bias expressed by gender-specific descriptions and inductions, and expressed gender-stereotyped attitudes, norms and beliefs. 
As a result, gender-swapped methods may generate nonsensical sentences under certain circumstances.

We also use the basic mitigation annotation patterns (Fig.~\ref{fig:CaseStudyEditing}).
These three major mitigation annotation patterns are not used exclusively in the annotation but optionally in combination. Except for the three mentioned patterns, we apply several other linguistic skills, including deleting gender-specific pronouns and replacing vehicles in gender-related metaphors, to mitigate the gender bias while keeping semantic information unchanged.
\end{document}

%% file: fig/fitler_pipeline.tex
\begin{figure}[!hbt]
\centering

\includegraphics[width=0.75\linewidth]{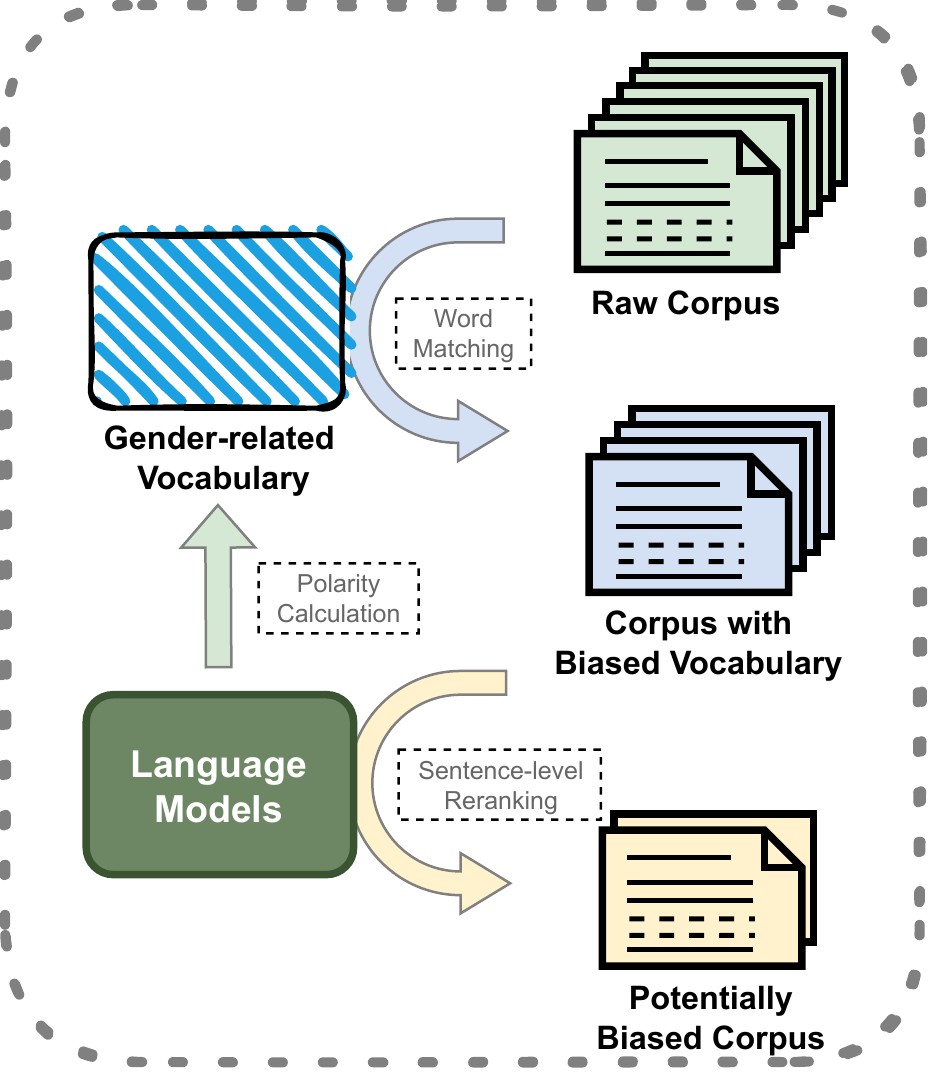} 
\vspace{2mm}
\caption{Pipeline of Retrieving and Filtering Potentially Biased Sentences from Raw Corpus for Human Annotation. }
\vspace{2mm}
\label{retriving_pipeline}
\end{figure}

%% file: main.bbl
\begin{thebibliography}{30}
\expandafter\ifx\csname natexlab\endcsname\relax\def\natexlab#1{#1}\fi

\bibitem[{Agarwal and Lavie(2007)}]{agarwal2007meteor}
Abhaya Agarwal and Alon Lavie. 2007.
\newblock Meteor: An automatic metric for mt evaluation with high levels of
  correlation with human judgments.
\newblock \emph{Proceedings of WMT-08}.

\bibitem[{Blodgett et~al.(2020)Blodgett, Barocas, Daumé~III, and
  Wallach}]{blodgett2020language}
Su~Lin Blodgett, Solon Barocas, Hal Daumé~III, and Hanna Wallach. 2020.
\newblock \href
  {https://www.microsoft.com/en-us/research/publication/language-technology-is-power-a-critical-survey-of-bias-in-nlp/}
  {Language (technology) is power: A critical survey of “bias” in nlp}.
\newblock In \emph{ACL}.

\bibitem[{Bolukbasi et~al.(2016)Bolukbasi, Chang, Zou, Saligrama, and
  Kalai}]{Bolukbasi2016ManIT}
Tolga Bolukbasi, Kai-Wei Chang, James~Y. Zou, Venkatesh Saligrama, and
  Adam~Tauman Kalai. 2016.
\newblock Man is to computer programmer as woman is to homemaker? debiasing
  word embeddings.
\newblock In \emph{NIPS}.

\bibitem[{Brown et~al.(2020)Brown, Mann, Ryder, Subbiah, Kaplan, Dhariwal,
  Neelakantan, Shyam, Sastry, Askell et~al.}]{brown2020language}
Tom Brown, Benjamin Mann, Nick Ryder, Melanie Subbiah, Jared~D Kaplan, Prafulla
  Dhariwal, Arvind Neelakantan, Pranav Shyam, Girish Sastry, Amanda Askell,
  et~al. 2020.
\newblock Language models are few-shot learners.
\newblock \emph{Advances in neural information processing systems},
  33:1877--1901.

\bibitem[{Chiril et~al.(2021)Chiril, Benamara, and Moriceau}]{chiril2021nice}
Patricia Chiril, Farah Benamara, and V{\'e}ronique Moriceau. 2021.
\newblock “be nice to your wife! the restaurants are closed”: Can gender
  stereotype detection improve sexism classification?
\newblock In \emph{Findings of the Association for Computational Linguistics:
  EMNLP 2021}, pages 2833--2844.

\bibitem[{Chiril et~al.(2020)Chiril, Moriceau, Benamara, Mari, Origgi, and
  Coulomb-Gully}]{chiril2020annotated}
Patricia Chiril, V{\'e}ronique Moriceau, Farah Benamara, Alda Mari, Gloria
  Origgi, and Marl{\`e}ne Coulomb-Gully. 2020.
\newblock An annotated corpus for sexism detection in french tweets.
\newblock In \emph{Proceedings of the 12th language resources and evaluation
  conference}, pages 1397--1403.

\bibitem[{Costa-juss{\`a}(2019)}]{costa2019analysis}
Marta~R Costa-juss{\`a}. 2019.
\newblock An analysis of gender bias studies in natural language processing.
\newblock \emph{Nature Machine Intelligence}, 1(11):495--496.

\bibitem[{Cui et~al.(2020)Cui, Che, Liu, Qin, Wang, and Hu}]{cui2020revisiting}
Yiming Cui, Wanxiang Che, Ting Liu, Bing Qin, Shijin Wang, and Guoping Hu.
  2020.
\newblock Revisiting pre-trained models for chinese natural language
  processing.
\newblock \emph{arXiv preprint arXiv:2004.13922}.

\bibitem[{Jiang et~al.(2022)Jiang, Yang, Liu, and Zubiaga}]{jiang2022swsr}
Aiqi Jiang, Xiaohan Yang, Yang Liu, and Arkaitz Zubiaga. 2022.
\newblock Swsr: A chinese dataset and lexicon for online sexism detection.
\newblock \emph{Online Social Networks and Media}, 27:100182.

\bibitem[{Jiao and Luo(2021)}]{jiao2021gender}
Meichun Jiao and Ziyang Luo. 2021.
\newblock Gender bias hidden behind chinese word embeddings: The case of
  chinese adjectives.
\newblock In \emph{Proceedings of the 3rd Workshop on Gender Bias in Natural
  Language Processing}, pages 8--15.

\bibitem[{King et~al.(2021)King, Scovelle, Meehl, Milner, and
  Priest}]{king2021gender}
Tania~L King, Anna~J Scovelle, Anneke Meehl, Allison~J Milner, and Naomi
  Priest. 2021.
\newblock Gender stereotypes and biases in early childhood: A systematic
  review.
\newblock \emph{Australasian Journal of Early Childhood}, 46(2):112--125.

\bibitem[{Lin(2004)}]{lin2004rouge}
Chin-Yew Lin. 2004.
\newblock Rouge: A package for automatic evaluation of summaries.
\newblock In \emph{Text summarization branches out}, pages 74--81.

\bibitem[{Lu et~al.(2020)Lu, Mardziel, Wu, Amancharla, and
  Datta}]{lu2020gender}
Kaiji Lu, Piotr Mardziel, Fangjing Wu, Preetam Amancharla, and Anupam Datta.
  2020.
\newblock Gender bias in neural natural language processing.
\newblock In \emph{Logic, Language, and Security}, pages 189--202. Springer.

\bibitem[{Papineni et~al.(2002)Papineni, Roukos, Ward, and
  Zhu}]{papineni2002bleu}
Kishore Papineni, Salim Roukos, Todd Ward, and Wei-Jing Zhu. 2002.
\newblock Bleu: a method for automatic evaluation of machine translation.
\newblock In \emph{Proceedings of the 40th annual meeting of the Association
  for Computational Linguistics}, pages 311--318.

\bibitem[{Parikh et~al.(2019)Parikh, Abburi, Badjatiya, Krishnan, Chhaya,
  Gupta, and Varma}]{parikh2019multi}
Pulkit Parikh, Harika Abburi, Pinkesh Badjatiya, Radhika Krishnan, Niyati
  Chhaya, Manish Gupta, and Vasudeva Varma. 2019.
\newblock Multi-label categorization of accounts of sexism using a neural
  framework.
\newblock \emph{arXiv preprint arXiv:1910.04602}.

\bibitem[{Qiu et~al.(2018)Qiu, Li, Li, Jiang, Hu, and Yang}]{qiu2018revisiting}
Yuanyuan Qiu, Hongzheng Li, Shen Li, Yingdi Jiang, Renfen Hu, and Lijiao Yang.
  2018.
\newblock Revisiting correlations between intrinsic and extrinsic evaluations
  of word embeddings.
\newblock In \emph{Chinese Computational Linguistics and Natural Language
  Processing Based on Naturally Annotated Big Data}, pages 209--221. Springer.

\bibitem[{Rudinger et~al.(2018)Rudinger, Naradowsky, Leonard, and
  Van~Durme}]{rudinger2018gender}
Rachel Rudinger, Jason Naradowsky, Brian Leonard, and Benjamin Van~Durme. 2018.
\newblock Gender bias in coreference resolution.
\newblock \emph{arXiv preprint arXiv:1804.09301}.

\bibitem[{Sahai and Sharma(2021)}]{sahai2021predicting}
Saumya Sahai and Dravyansh Sharma. 2021.
\newblock Predicting and explaining french grammatical gender.
\newblock In \emph{Proceedings of the Third Workshop on Computational Typology
  and Multilingual NLP}, pages 90--96.

\bibitem[{Srivastava et~al.(2022)Srivastava, Rastogi, Rao, Shoeb, Abid, Fisch,
  Brown, Santoro, Gupta, Garriga-Alonso et~al.}]{srivastava2022beyond}
Aarohi Srivastava, Abhinav Rastogi, Abhishek Rao, Abu Awal~Md Shoeb, Abubakar
  Abid, Adam Fisch, Adam~R Brown, Adam Santoro, Aditya Gupta, Adri{\`a}
  Garriga-Alonso, et~al. 2022.
\newblock Beyond the imitation game: Quantifying and extrapolating the
  capabilities of language models.
\newblock \emph{arXiv preprint arXiv:2206.04615}.

\bibitem[{Sun et~al.(2019)Sun, Gaut, Tang, Huang, ElSherief, Zhao, Mirza,
  Belding, Chang, and Wang}]{sun2019mitigating}
Tony Sun, Andrew Gaut, Shirlyn Tang, Yuxin Huang, Mai ElSherief, Jieyu Zhao,
  Diba Mirza, Elizabeth Belding, Kai-Wei Chang, and William~Yang Wang. 2019.
\newblock Mitigating gender bias in natural language processing: Literature
  review.
\newblock \emph{arXiv preprint arXiv:1906.08976}.

\bibitem[{Waseem(2016)}]{waseem-2016-racist}
Zeerak Waseem. 2016.
\newblock \href {https://doi.org/10.18653/v1/W16-5618} {Are you a racist or am
  {I} seeing things? annotator influence on hate speech detection on
  {T}witter}.
\newblock In \emph{Proceedings of the First Workshop on {NLP} and Computational
  Social Science}, pages 138--142, Austin, Texas. Association for Computational
  Linguistics.

\bibitem[{Webster et~al.(2018)Webster, Recasens, Axelrod, and
  Baldridge}]{webster2018mind}
Kellie Webster, Marta Recasens, Vera Axelrod, and Jason Baldridge. 2018.
\newblock Mind the gap: A balanced corpus of gendered ambiguous pronouns.
\newblock \emph{Transactions of the Association for Computational Linguistics},
  6:605--617.

\bibitem[{Weidong et~al.(2019)Weidong, Rui, Baobao, Yirong, and
  Chen}]{zhan2019CCL}
Zhan Weidong, Guo Rui, Chang Baobao, Chen Yirong, and Long Chen. 2019.
\newblock Development of {P}eking {U}niversity {CCL} {C}orpus.
\newblock In \emph{Chinese {C}orpus {L}inguistic {J}ournal}.

\bibitem[{Xu et~al.(2019)Xu, Zhang, Wu, and Wang}]{xu2019cinderella}
Huimin Xu, Zhang Zhang, Lingfei Wu, and Cheng-Jun Wang. 2019.
\newblock The cinderella complex: Word embeddings reveal gender stereotypes in
  movies and books.
\newblock \emph{PloS one}, 14(11):e0225385.

\bibitem[{Zhang et~al.(2019)Zhang, Han, Liu, Jiang, Sun, and
  Liu}]{zhang2019ernie}
Zhengyan Zhang, Xu~Han, Zhiyuan Liu, Xin Jiang, Maosong Sun, and Qun Liu. 2019.
\newblock Ernie: Enhanced language representation with informative entities.
\newblock \emph{arXiv preprint arXiv:1905.07129}.

\bibitem[{Zhao et~al.(2017)Zhao, Wang, Yatskar, Ordonez, and
  Chang}]{zhao2017men}
Jieyu Zhao, Tianlu Wang, Mark Yatskar, Vicente Ordonez, and Kai-Wei Chang.
  2017.
\newblock Men also like shopping: Reducing gender bias amplification using
  corpus-level constraints.
\newblock \emph{arXiv preprint arXiv:1707.09457}.

\bibitem[{Zhao et~al.(2018)Zhao, Zhou, Li, Wang, and Chang}]{zhao2018learning}
Jieyu Zhao, Yichao Zhou, Zeyu Li, Wei Wang, and Kai-Wei Chang. 2018.
\newblock Learning gender-neutral word embeddings.
\newblock \emph{arXiv preprint arXiv:1809.01496}.

\bibitem[{Zhao et~al.(2021)Zhao, Du, Zhu, and Liu}]{zhao2021}
Jishun Zhao, Bingjie Du, Shucheng Zhu, and Pengyuan Liu. 2021.
\newblock Construction of chinese sentence-level gender-unbiased data set and
  evaluation of gender bias in pre-training language.
\newblock In \emph{Proceedings of the 20th Chinese National Conference on
  Computational Linguistics}, pages 564--575.

\bibitem[{Zhou et~al.(2019)Zhou, Shi, Zhao, Huang, Chen, Cotterell, and
  Chang}]{zhou2019examining}
Pei Zhou, Weijia Shi, Jieyu Zhao, Kuan-Hao Huang, Muhao Chen, Ryan Cotterell,
  and Kai-Wei Chang. 2019.
\newblock Examining gender bias in languages with grammatical gender.
\newblock \emph{arXiv preprint arXiv:1909.02224}.

\bibitem[{Zhu and Liu(2020)}]{zhu2020great}
Shucheng Zhu and Pengyuan Liu. 2020.
\newblock Great males and stubborn females: A diachronic study of corpus-based
  gendered skewness in chinese adjectives.
\newblock In \emph{Proceedings of the 19th Chinese National Conference on
  Computational Linguistics}, pages 31--42.

\end{thebibliography}
